%% file: article.tex
\documentclass{article}
\usepackage{natbib}
\usepackage[title]{appendix}
\newcommand{\Ex}{\mathbb{E}}
\usepackage{graphicx} 
\usepackage{hyperref}
\hypersetup{colorlinks,
            linktoc=page,
            urlcolor=red,
            linkcolor=red,
            citecolor=blue}

\usepackage[most]{tcolorbox}
\usepackage{amsfonts}
\usepackage{amsmath}
\usepackage{amssymb}
\usepackage{setspace}
\usepackage{authblk}

\DeclareMathOperator*{\argmax}{arg\,max}
\DeclareMathOperator*{\argmin}{arg\,min}

\usepackage{color}
\definecolor{deepblue}{rgb}{0,0,0.5}
\definecolor{deepred}{rgb}{0.6,0,0}
\definecolor{deepgreen}{rgb}{0,0.5,0}

\usepackage{listings}
\usepackage{courier}

\newcommand\pythonstyle{\lstset{
language=Python,
basicstyle=\footnotesize\ttfamily,
breaklines=true
morekeywords={self},              
keywordstyle=\ttfamily\color{deepblue},
emph={MyClass,__init__},          
emphstyle=\ttfamily\color{deepred},    
stringstyle=\color{deepgreen},
frame=tb,                         
showstringspaces=false
}}

\lstnewenvironment{python}[1][]
{
\pythonstyle
\lstset{#1}
}
{}

\advance\textheight by \topskip
\tcbset{boxstyle/.style={width=\textwidth,breakable,colback=red!5!white,colframe=red!75!red,fonttitle=\bfseries,code={\singlespacing},before upper={\parindent15pt}}}

\newtcolorbox[auto counter,list inside=boxstyle]{TCBBox}[2][]{boxstyle,title=Box~\thetcbcounter: #2, #1}

\newcommand{\listofboxes}{%
  \tcblistof[\section*]{boxstyle}{Boxes}
}

\title{A Primer on Deep Learning for Causal Inference}
\author[1,2,3]{Bernard J. Koch\thanks{bernardkoch@ucla.edu}}
\author[4]{Tim Sainburg}
\author[1]{Pablo Geraldo Bast\'ias}
\author[5]{Song Jiang}
\author[5]{Yizhou Sun}
\author[1,6]{Jacob Foster}
\affil[1]{UCLA Department of Sociology}
\affil[2]{Northwestern Kellogg School of Management}
\affil[3]{University of Chicago Department of Sociology}
\affil[4]{Harvard Medical School}
\affil[5]{UCLA Department of Computer Science}
\affil[6]{Santa Fe Institute}
\date{April 2023}

\begin{document}

\maketitle

\begin{abstract}
This primer systematizes the emerging literature on causal inference using deep neural networks under the potential outcomes framework. It provides an intuitive introduction on building and optimizing custom deep learning models and shows how to adapt them to estimate/predict heterogeneous treatment effects. It also discusses ongoing work to extend causal inference to settings where confounding is non-linear, time-varying, or encoded in text, networks, and images. To maximize accessibility, we also introduce prerequisite concepts from causal inference and deep learning. The primer differs from other treatments of deep learning and causal inference in its sharp focus on observational causal estimation, its extended exposition of key algorithms, and its \href{https://github.com/kochbj/Deep-Learning-for-Causal-Inference}{detailed tutorials} for implementing, training, and selecting among deep estimators in Tensorflow 2 and PyTorch.
\end{abstract}

\newpage
\singlespacing
\tableofcontents

\listofboxes

\newpage
\doublespacing

\input{content/introduction.tex}
\input{content/deeplearning.tex}

\input{content/causalinference.tex}
\input{content/architectures.tex}
\input{content/outcomemodeling.tex}
\input{content/balancing.tex}
\input{content/propensity.tex}

\input{content/interpretation.tex}
\input{content/nontraditional.tex}
\input{content/conclusion.tex}

\bibliographystyle{asr}
\bibliography{article}

\begin{appendices}
\input{content/appendix.tex}
\end{appendices}
\end{document}

%% file: content/introduction.tex
\section{Introduction}
\label{section:introduction}

This primer aims to introduce social science readers to an exciting literature exploring how deep neural networks can be used to estimate causal effects. In recent years, both causal inference frameworks and deep learning have seen rapid adoption across science, industry, and medicine. Causal inference has a long tradition in the social sciences, and social scientists are increasingly exploring the use of machine learning for causal inference \citep{Athey2016,wager2018, chernozhukov2016double}. Nevertheless, deep learning remains conspicuously underutilized by social scientists compared to other ML approaches, both for causal inference and more generally.

The deep learning revolution has been spurred by the flexibility and expressiveness of these models. Neural networks are nearly non-parametric and can theoretically approximate any continuous function \citep{Cybenko1989ApproximationBS}, making them well suited for both classification and regression tasks. Furthermore, they can be configured with different architectures and objectives to learn from a variety of quantitative data as well as text, images, video, networks, and speech. These advantages allow them to learn vector ``representations" of complex data with emergent properties. Simple examples of representation learning include the Word2Vec algorithm that discovers semantic relationship between words in texts, or face classification models that learn vectors describing facial features \citep{mikolov2013}. More recently, generative models like DALL-E, Stable Diffusion, and ChatGPT have shown how coherent text passages and life-like images can be reconstructed from learned representations. 

Here we explore the potential for leveraging these advantages to estimate causal effects. Causal inference frameworks are non-parametric, but the linear models traditionally used to estimate causal effects require strong parametric assumptions. In contrast, the nearly non-parametric nature of neural networks allows us to estimate smooth response surfaces that capture heterogeneous treatment effects for individual units with low bias.\footnote{Neural networks can have hundreds to billions of parameters making them effectively non-parametric. The risks of overparameterization of neural networks are discussed in Section \ref{section:primerdl}.} The ability of these models to learn from complex data means we can extend causal inference to new settings where confounding is complicated, time-varying, or even encoded in texts, graphs, or images (see Box \ref{Box:nontradex} for hypothetical examples). Lastly given the right objectives, neural networks promise to learn deconfounded representations of data, presenting a new strategy for treatment modeling.

This primer synthesizes existing literature on deep causal estimators, but it is not a review; its goals are fundamentally pedagogical and prospective rather than retrospective. In Section \ref{section:primerdl}, we introduce social scientists to the fundamental concepts of deep learning, and the basic workflow for building and training their own deep neural networks within a supervised learning framework. For readers unfamiliar with causal inference, Section \ref{section:CI} introduces the assumptions of causal identification and three fundamental estimation strategies within the selection on observables design: matching, outcome modeling, and inverse propensity score weighting. Machine learning models often perform poorly in both theory and practice when only one of these strategies is employed, so we also introduce the concept of double robustness.

Section \ref{section:mainbody} is the main body of the article. Here we introduce four related deep learning models for the estimation of heterogeneous treatment effects: the S-learner, T-learner, TARNet and Dragonnet \citep{Shalit2017,Shi2019}. Although this literature is rapidly evolving, these four models are sufficient to illustrate how traditional estimation strategies can be used in creative ways that leverage the key strengths of neural networks (i.e., deconfounding through representation learning, semi-parametric inference). Section \ref{section:interpretation} deals with the practical considerations of building confidence intervals and interpreting neural networks. These guidelines are concretized in the \href{https://github.com/kochbj/Deep-Learning-for-Causal-Inference}{companion online tutorials}, which show readers how to implement and interpret the models described in Section \ref{section:mainbody} in Tensorflow 2 and PyTorch.

In Section \ref{section:nontraditional}, we focus on the future of deep causal information: estimators that can disentange counfounding relationships embedded within texts, images, graphs, or time-varying data. In the interest of clarity, we give hypothetical examples of the types of questions social scientists might answer with these models, and briefly describe ongoing research on each of these modalities. For fuller treatments of some of these models, see the appendix. We conclude with a discussion of how neural networks fit into the broader literature on machine learning for causal inference (Section \ref{section:conclusions}).

The primer makes multiple contributions. First, it is one of the first pieces in the sociological literature to introduce the fundamentals of deep learning not only at a conceptual level (e.g, backpropagation, representation learning), but at a practical one (e.g., validation, hyperparameter tuning). Our recommendations for training and interpreting neural networks are supported by heavily annotated tutorials that teach readers without prior familiarity with deep learning how to build their own custom models in Tensorflow 2 and PyTorch. Second, we use this foundation and select examples to build intuition on how the core strengths of deep learning can be leveraged for causal inference. Finally, we highlight future directions for this literature and argue why the future of causal estimation runs through deep learning.

\begin{TCBBox}[label=Box:nontradex]{Example Scenarios for Causal Inference with Non-Traditional Data}
\textbf{Text.} As a motivating example, \citet{Veitch2019UsingTE} consider the effect of the author's reported gender ($T$) on the number of upvotes a Reddit post receives ($Y$). However gender may also ``affect the text of the post, e.g., through tone, style, or topic
choices, which also affects its score [($X$)]." Controlling for a representation of the text would allow the analyst to more accurately estimate the direct effect of gender.

\textbf{Images.} \citet{todorov2005inferences} showed that split second-judgments of a politician's competence ($T$) from pictures ($X$) of their face is predictive of their electability ($Y$). When attempting to replicate this study using machine learning classifiers rather than human classifiers, \citet{joo2015automated} suggest that the age of the face ($Z$) is a not-so-obvious confounder: while older individuals are more likely to appear competent, they are also more likely to be incumbents. Even if age is unknown, using neural networks to control for confounders implicitly encoded in the image (like age) could reduce bias.

\textbf{Networks.} \citet{nagpal2020} explore the question of which types of prescription opioids (e.g., natural, semi-synthetic, synthetic) ($T$) are most likely to cause long term addiction ($Y$). Because of predisposition to different injuries, type of employment ($X$) could be a common cause of both treatment and outcome. Suppose job type is unobserved, but we know that patients are likely to associate with coworkers through homophily. To capture some of the effects of this latent unobserved confounder, analysts might choose to control for a representation of the patient's position in their social network when estimating the causal effect.

\end{TCBBox}

%% file: content/deeplearning.tex
\section{Deep Learning Fundamentals}
\label{section:primerdl}

\subsection{Artificial Neural Networks}

\begin{TCBBox}[label=Box:supervisedlearning]{Basic Introduction to Supervised Learning}

Deep learning algorithms have most commonly been adapted for causal inference using supervised machine learning, the most popular learning framework within the field.\footnote{The other two prominent paradigms are unsupervised learning and reinforcement learning.} The goal of supervised learning is teach a model a non-linear function that transforms covariates/features $X$ into predicted outcomes $\hat{Y}$ in unseen data. The model learns this function from labeled examples of $X_\text{tr}$ and $Y_{tr}$ in a \textbf{training dataset}.

As in traditional statistical analyses, the function is learned by optimizing the model's parameters such that they minimize the error between its predictions $\hat{Y_{tr}}$ and the true values $Y_{tr}$ using a \textbf{loss function} (e.g., a likelihood). In a traditional social science analysis focused on inference, we would stop here and interpret these parameters. In supervised machine learning where the focus is on generalization to unseen data, the model is ultimately used to predict outcomes $Y_{te}$ in a \textbf{test dataset} of previously unseen covariates/features $X_{te}$. This framework can be generically applied to cases where $Y$ is categorical (called classification problems), and where $Y$ is continuous (called regression problems).

Statistical learning theory articulates the central challenge of supervised learning as a balance between \textbf{overfitting} and \textbf{underfitting} the training dataset. This is also called the \textbf{bias-variance" tradeoff}. In a regression context, bias error is the difference between the expected value of Y and the expected value of the mapping function learned by the model.\footnote{Note that bias in statistical learning theory is not equivalent to bias of a statistical estimator.} High bias typically results from an algorithm that has not sufficiently learned the relationships in the training dataset (i.e., underfit the data). In contrast, an algorithm that has learned the training dataset so closely that it is fitting noise in the sample (i.e., overfitting) is likely to generalize poorly, producing out-of-sample predictions with high variance. Underfitting can be easily diagnosed and addressed by increasing the complexity of the model. In the case of deep learning, model complexity can be increased by adding additional layers or parameters/neurons.

Diagnosing and addressing overfitting is a more challenging problem. In supervised learning, overfitting is diagnosed after training (but before testing) by assessing predictive performance in a reserved portion of the training set called the \textbf{validation set}. If the model fits the training dataset well but performs poorly in the validation set, it is likely to generalize poorly to the test set as well. To prevent overfitting, \textbf{regularization} techniques can be used to simplify the complexity of the model. Training and regularization of neural networks is discussed in detail in Section \ref{section:practice}. For a full treatment of supervised learning and statistical learning theory, see \citet{hastie2009elements}.
\end{TCBBox}

\begin{figure}
    \centering
    \includegraphics[width=\linewidth]{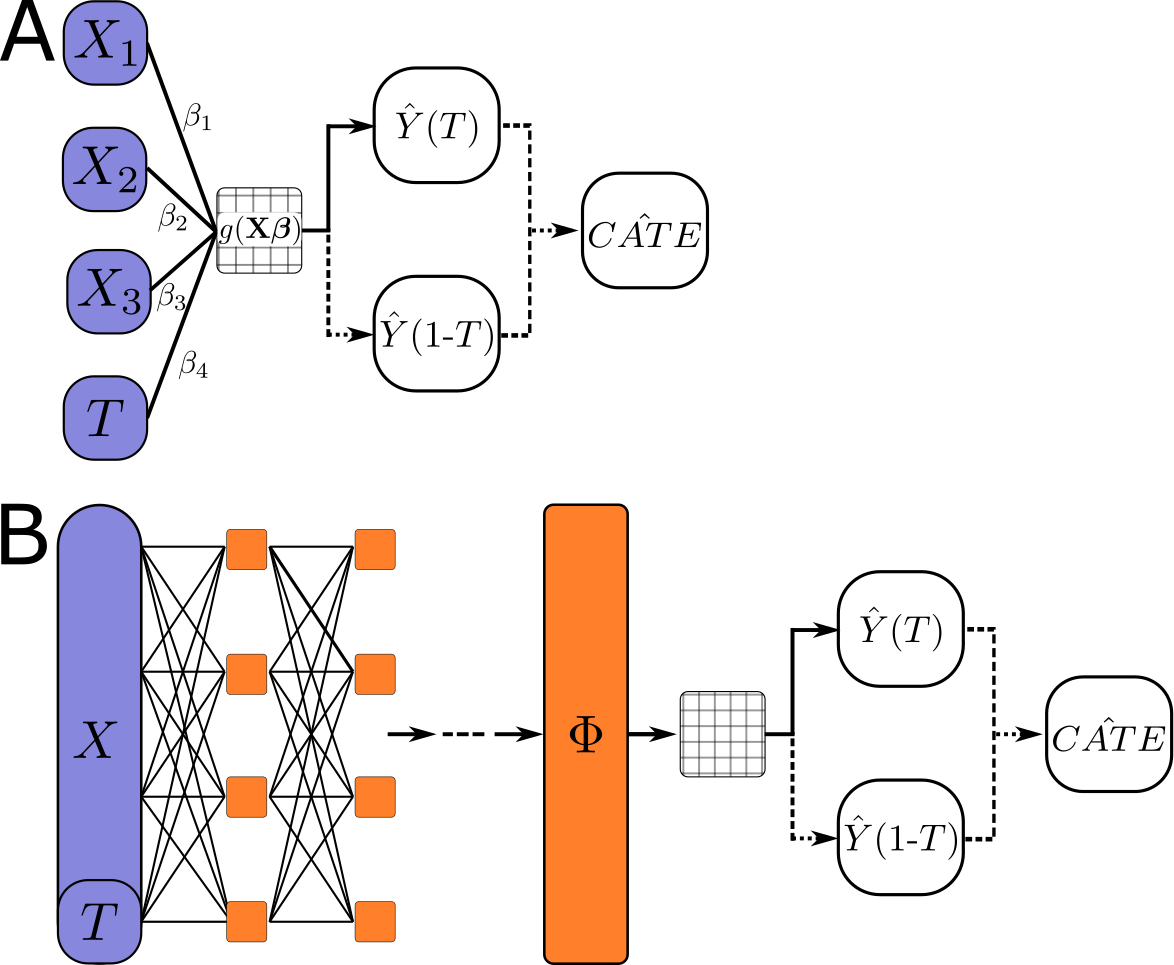}
    \caption{\textbf{A: Generalized linear model represented as a computational graph.} Observable covariates $X_1,X_2,X_3$ and treatment status $T$ depicted in rounded purple boxes. Each of the lines between the  rounded purple inputs and the textured box represents a parameter (i.e., a $\beta$ in a generalized linear model equation). The textured box is an ``output neuron" that sums it's weighted inputs, performs a transformation $g$ (the link function in GLM; in this case the identity function), and predicts the conditional outcome $\hat{Y}(T)$. Instead of theoretically interpreting these parameters from an inferential statistics perspective, machine learning approaches typically use the predicted observed and unobserved potential outcomes for plug-in estimation of causal estimands (e.g., the conditional average treatment effect $\hat{CATE}$). After training, setting $T$ to $1-T$ for each observation can predict the unobserved potential outcome $\hat{Y}(1-T)$. Because this operation occurs after prediction and does not feed a gradient back to the network to optimize the parameters, it is depicted here with a dotted line. Plug-in calculation of $\hat{CATE}$ is similarly shown with a dotted line.
    \\
    \textbf{B: Feed-forward neural network (S-learner).} In a feed-forward neural network, additional fully connected (parameterized) layers of neurons are added between the inputs (rounded purple) and output neuron. The size of the input covariates and hidden layers are generically abstracted as boxes (orange). The final hidden layer before the output neuron is denoted $\Phi$ because the hidden layers collectively encode a representation function (see section \ref{section:replearning}). In causal inference settings, this architecture is sometimes called a S(ingle)-learner because one feed-forward network learns to predict both potential outcomes.}
    \label{fig:perceptron}
\end{figure}

Artificial neural networks (ANN) are statistical models inspired by the human brain \citep{brand_koch_xu,Goodfellow-et-al-2016}. In an ANN, each ``neuron" in the network takes the weighted sum of its inputs (the outputs of other neurons) and transforms them using a differentiable, non-linear function (e.g. sigmoid, rectified linear unit) that outputs a value between 0 and 1 if the transformed value is above some threshold. Neurons are arrayed in layers where an input layer takes the raw data, and neurons in subsequent layers take the weighted sum of outputs in previous layers as input. An ``output" layer contains a neuron for each of the predicted outcomes with transformation functions appropriate to those outcomes. For example, a regression network that predicts one outcome will have a single output neuron without a transformation function so that it produces a real number. A regression network without any hidden layers corresponds exactly to a generalized linear model (Fig. \ref{fig:perceptron}A). When additional ``hidden" layers are added between the input and output layers, the architecture is called a \textbf{feed-forward network} or \textbf{multi-layer perceptron} (Fig. \ref{fig:perceptron}B). A neural network with multiple hidden layers is called a ``deep" network, hence the name ``deep learning" \citep{lecun2015deep}. A neural network with a single, large enough hidden layer can theoretically approximate any continuous function \citep{Cybenko1989ApproximationBS}. 

\begin{TCBBox}[label=Box:training]{Reading Machine Learning Papers: Computational Graphs and Loss Functions}
Within the machine learning literature, novel algorithms are often presented in terms of their computational graph and loss function. A computational graph (not to be confused with a causal graph) uses arrows to depicts the flow of data from the inputs of a neural network,  through parameters, to the outputs. Layers of neurons or specialized sub-architectures are often generically abstracted as shapes. In our diagrams, we use rounded purple shapes to represent observables, orange rectangles for representation layers of the network, rounded white shapes for produced outputs, and textured rectangles for outcome modeling layers. Operations that are computed \textit{after} prediction (i.e., for which an error gradient is not calculated) are shown with dashed lines (e.g., plug-in estimation of causal estimands).

Along with the architecture, the loss function of a neural network is the primary means for the analyst to dictate what types of representations a neural network learns and what types of outputs it produces. In multi-task learning settings, we denote joint loss functions for an entire network as a weighted sum of the losses for substituent tasks and modules. These specific losses are weighted by hyperparameters. For example, we might weight the joint loss for a network that predicts outcomes and propensity scores as:

\begin{equation*}
    \argmin_{h,\pi}\mathcal{L} = \mathcal{L}_h + \lambda \mathcal{L}_\pi = MSE(Y,h(X,T)) + \lambda BCE(T,\pi(X,T))
\end{equation*}
where $h(X,T)$ is the predicted potential outcome, $\pi(X,T)$ is the predicted propensity score, $\lambda$ is a hyperparameter and MSE and BCE stand for mean squared error and binary cross entropy (i.e., log loss), common losses for regression and binary classification respectively (Box \ref{Box:Notation}). 
\end{TCBBox}

Neural networks are trained to predict their outcomes by optimizing a \textbf{loss function} (also called an objective or cost function). During training, the \textbf{backpropagation} algorithm
uses calculus's chain rule to assign portions of the total error in the loss function to each neuron in the network. An optimizer, such as the stochastic gradient descent algorithm or the currently popular ADAM algorithm \citep{Kingma2014AdamAM}, then moves each parameter in the opposite direction of this error gradient. Neural networks first rose to popularity in the 1980s but fell out of favor compared to other machine learning model families (e.g., support vector machines) due to their expense of training. By the late 2000s, improvements to backpropagation, advances in computing power (i.e., graphic cards), and access to larger datasets collectively enabled a deep learning revolution where ANNs began to significantly outperform other model families. Today, deep learning is the hegemonic machine learning approach in industries and fields other than social science.

\subsection{Deep Learning in Practice}

\label{section:practice}
\begin{figure}
    \centering
    \includegraphics[width=\linewidth]{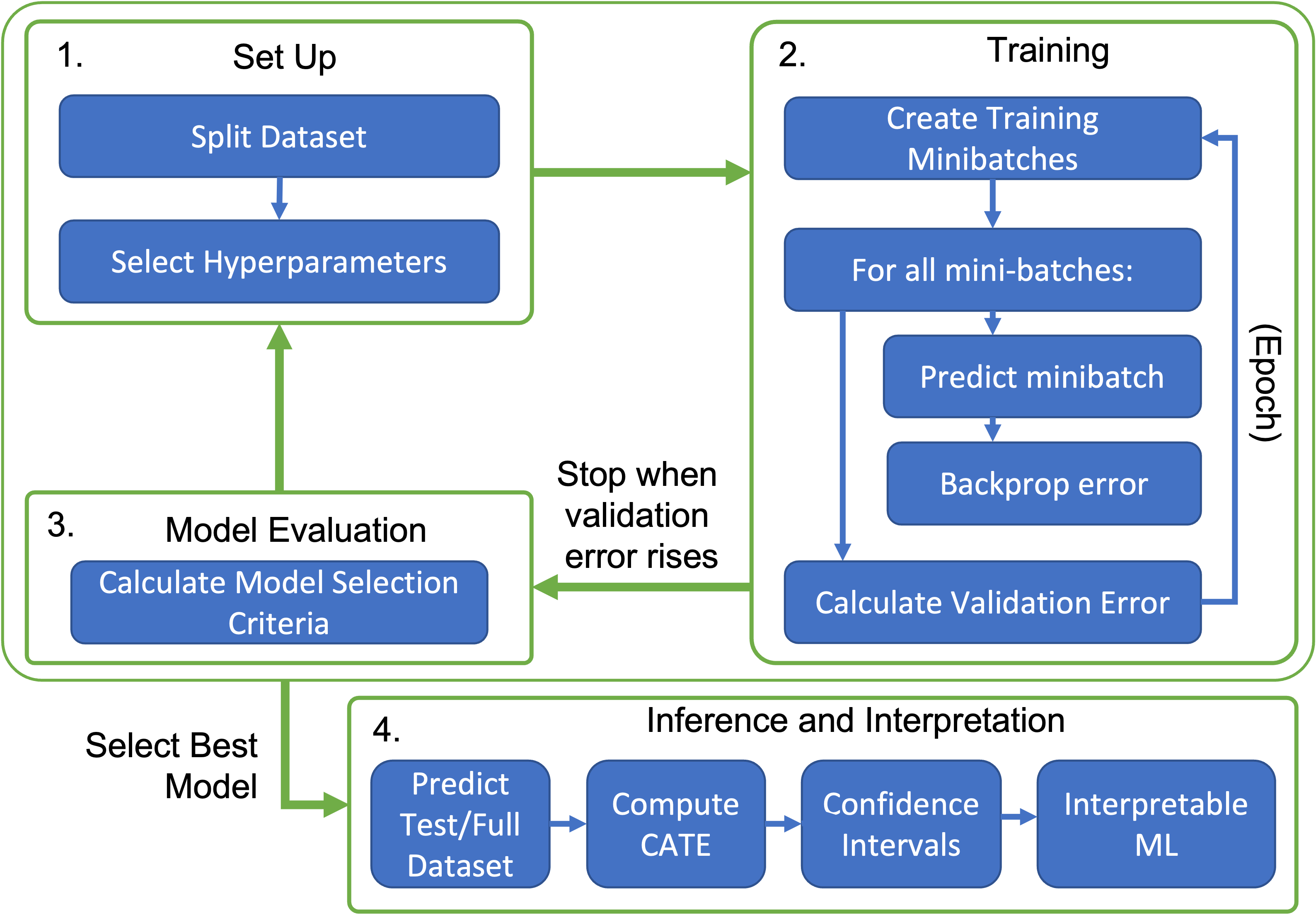}
    \caption{\textbf{Supervised Deep Learning Workflow.} \textbf{1) Set Up:} The first step in training a deep learning model is splitting the data into a training set, validation set, and optionally a test set. Initial hyperparameters are then selected from a set of choices specified by the user. \textbf{2) Training:} In each iteration of the training process (called an epoch), the training set is randomly divided into small minibatches For each minibatch, the network makes predictions for all units, and calculates the error gradients to be assigned to each neuron in the network based on those predictions. An optimizer then move the network's parameters in the opposite direction of the error gradient. After all minibatches have been trained (one epoch), error is calculated on the entire validation set. This whole process is repeated up until the validation error stops decreasing (to avoid overfitting). \textbf{3) Model Evaluation:} A criterion (typically the validation error) is used to evaluate the performance of this hyperparameterization. New hyperparameters are then selected using a hyperparameter optimization algorithm (eg. Grid search, Bayesian hyperparameter optimization, genetic algorithms) and steps 1 and 2 are repeated. Once the hyperparameter optimization algorithm has completed its search, the ``best" model is selected for inference. \textbf{4) Inference and interpretation:} With a model selected, the analyst is now ready to apply it to their test data (or in the case of statistical inference, potentially the full dataset). Predictions of the outcomes and/or propensity score can then be used to compute the CATE and calculate confidence intervals. Feature importance algorithms like SHAP or Integrated Gradients can also be used to interpret the CATE estimates.}
    \label{fig:workflow}
\end{figure}

This section focuses on the practice of training neural networks within a supervised learning framework. While the principles behind supervised machine learning are universal, the workflow for neural networks differs substantially from other ML approaches (e.g., random forests, support vector machines) in practice. Figure \ref{fig:workflow} presents this workflow in four different parts: Set Up, Training, Model Evaluation, and Interpretation. We delve into each of these topics in more detail below. Box \ref{Box:supervisedlearning} contains a basic introduction to supervised learning for unfamiliar readers.

\subsubsection{Set Up and Hyperparameters}

The first step in training a neural network, as in other types of supervised machine learning, is to split your dataset into training, validation, and testing datasets (Fig. \ref{fig:workflow}A). If the network is being used for statistical inference, as here, the testing dataset is optional, and inference may be conducted on just the validation set or the full dataset.

While the computational graph and loss function define a deep learning architecture (Box \ref{Box:training}), actual implementations can vary significantly due to the choice of hyperparameters. In supervised machine learning, \textbf{hyperparameters} are parameters that are not learned automatically when training the model, but must be specified by the analyst. In deep learning, architectural hyperparameters include the number of layers to use for each section of the computational graph, the number of neurons to use in each layer, and the activation functions to be used by neurons. While some basic rules of thumb apply (e.g., use fewer layers than neurons), these choices remain poorly understood theoretically\footnote{For some interesting work on understanding neural networks theoretically from a statistical physics perspective see \citet{PDLT-2022}.}; Decisions are generally made by comparing empirical performance on the validation set, a practice called \textit{hyperparameter tuning}.

\subsubsection{Training and Regularization}
\label{subsection:trainingregularization}

Neural networks are trained by repeatedly making predictions from the training set, calculating error gradients for each parameter, and backpropagating small fractions of those error gradients. (Fig. \ref{fig:workflow} B).  A full pass through examples in the training set is called a training loop or \textbf{epoch}. At the beginning of each epoch, the training set is divided into \textbf{mini-batches} of 2 to 1000 units, randomly sampled without replacement. This practice not only aids in memory management, it also improves optimization. Using small random samples reduces the risk of large “exploding” error gradients, particularly early in the training, that could cause the model to overshoot optimal solutions and instead get stuck in local minima.

The size of mini-batches can be considered a hyperparameter.\footnote{In the specific context of causal inference, we recommend not having mini-batches that are too small such that the model can learn from both treated and control units with sufficient overlap.} Because a mini-batch of data is only a sample of a sample (the training dataset), the optimizer only adjusts weight parameters by a fraction of the error gradient (the \textbf{learning rate}) to avoid overfitting. The learning rate is also a hyperparameter, that typically varies between 0.0001 and 0.01.

The non-convex nature of most loss functions\footnote{In convex functions (e.g. the OLS loss), there is a single minimum, so optimizing the function means that you will always converge at the same parameter weights. This is not the case for non-convex functions which may have many local minima.} means that optimization often requires hundreds to potentially millions of epochs of training. Moreover, neural networks are highly susceptible to overfitting because it is easy to overparameterize them with excessive neurons/layers. To ward against overfitting, error metrics on the complete validation set are computed at the end of every epoch. In a regularization practice called “\textbf{early stopping},” analysts usually stop training once validation metrics stop improving.
Other common regularization techniques include \textbf{weight decay} (i.e., $\ell^2$ norm, ridge, or Tikhonov) penalties on the parameters, dropout of neurons during training, and batch normalization. 

\textbf{Dropout} is a regularization technique in deep learning where certain nodes are randomly silenced from training during a given epoch \citep{Srivastava2014DropoutAS}. The general idea of dropout is to force two neurons in the same layer to learn different aspects of the covariate/feature space and reduce overfitting. \textbf{Batch normalization} is another regularization technique applied to a layer of neurons \citep{Ioffe2015BatchNA}. By standardizing (i.e. z-scoring) the inputs to a layer on a per-batch basis and then rescaling them using trainable parameters, batch normalization smooths the optimization of the loss function. The addition and extent of each of these regularization techniques can be treated as hyperparemeters.

\subsubsection{Model Selection}
(\textit{Tutorial 2 \href{https://colab.research.google.com/drive/1MBUkQO7rh89JrlAV0p1aLNoF9CIB3rZZ?usp=sharing}{\includegraphics[height=\fontcharht\font`\B,keepaspectratio]{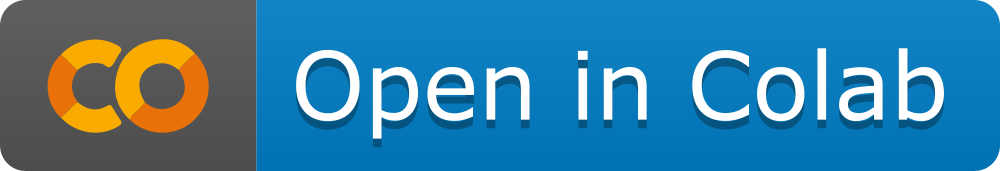} }})
\label{subsection:modelselection}

After the model has been trained, the analyst compares models assembled with different hyperparameterizations or initial parameter values (Fig. \ref{fig:workflow}C). Hyperparameterizations can be chosen using random search, an exhaustive grid search of all possible combinations, or strategic search algorithms like Bayesian hyperparameter optimization or evolutionary optimization \citep{snoek2012}. Validation loss metrics on the final epoch are commonly used for these comparisons. 

Model selection for causal estimators is complicated by the fundamental problem of causal inference: we are not actually interested in the observed “factual” outcomes and propensity scores, but the CATE and ATE. In the case of algorithms like Dragonnet \ref{section:ipw} where the validation loss explicitly targets a causal quantity, we use that as the model selection criterion. In cases where the algorithm is only trained for outcome modeling or propensity modeling, other solutions are needed. In the Appendix, we describe \citet{johannson2020}'s proposal to use matching on a nearest neighbor approximation of the \textit{Precision in Estimated Heterogeneous Effects} (PEHE), a measure of CATE bias, as an alternative model selection metric (Appendix A.\ref{appendix:peheselection}).

The development of more sophisticated methods for model selection of causal estimators through data simulation is an active area of research within this literature.\footnote{We note that crossfitting \citep{zivich2021machine}, another approach that has emerged for model selection of other types of machine learning causal estimators may work for the models discussed here, but is likely data-inefficient.} For example, \citet{parikh2022} use deep generative models to approximate the data generating distribution under weak, non-parametric assumptions. \citet{alaa2019validating} independently model each  outcome and the propensity score before using influence functions to assess model error.

\subsection{Representation Learning and Multitask Learning}

\label{section:replearning}
One comparative advantage of deep learning over other machine learning approaches has been the ability of ANNs to encode and automatically compress informative features from complex data into flexible, relevant ``\textbf{representations}" or ``embeddings" that make downstream supervised learning tasks easier \citep{Goodfellow-et-al-2016, bengio2013deep}. While other machine learning approaches may also encode representations, they often require extensive pre-processing to create useful features for the algorithm (i.e., feature engineering). Through the lens of representation learning, a geometric interpretation of the role of each layer in a supervised neural network is to transform its inputs (either raw data or output of previous layers) into a typically lower (but possibly higher) dimensional vector space. As a means to share statistical power, encoded representations can also be jointly learned for two tasks at once in \textbf{multi-task learning}. 

The simplest example of a representation might be the final layer in a feed-forward network, where the early layers of the network can be understood as non-linearly encoding the inputs into an array of latent linear features for the output neuron \citep{Goodfellow-et-al-2016} (Fig. \ref{fig:perceptron}B). A famous example of representation learning is the use of neural networks for face detection. Examining the representations produced by each layer of these networks shows that each subsequent layer seems to capture increasingly abstract features of a face (first edges, then noses and eyes, and finally whole faces) \citep{lecun2015deep}. A more familiar example of representation learning to social scientists might be word vector models like Word2Vec \citep{mikolov2013}. Word2Vec is a neural network with one hidden layer and one output layer where words that are semantically similar are closer together in the representation space created by the hidden layer of the network. 

\begin{figure}
    \centering
    \includegraphics[width=\linewidth]{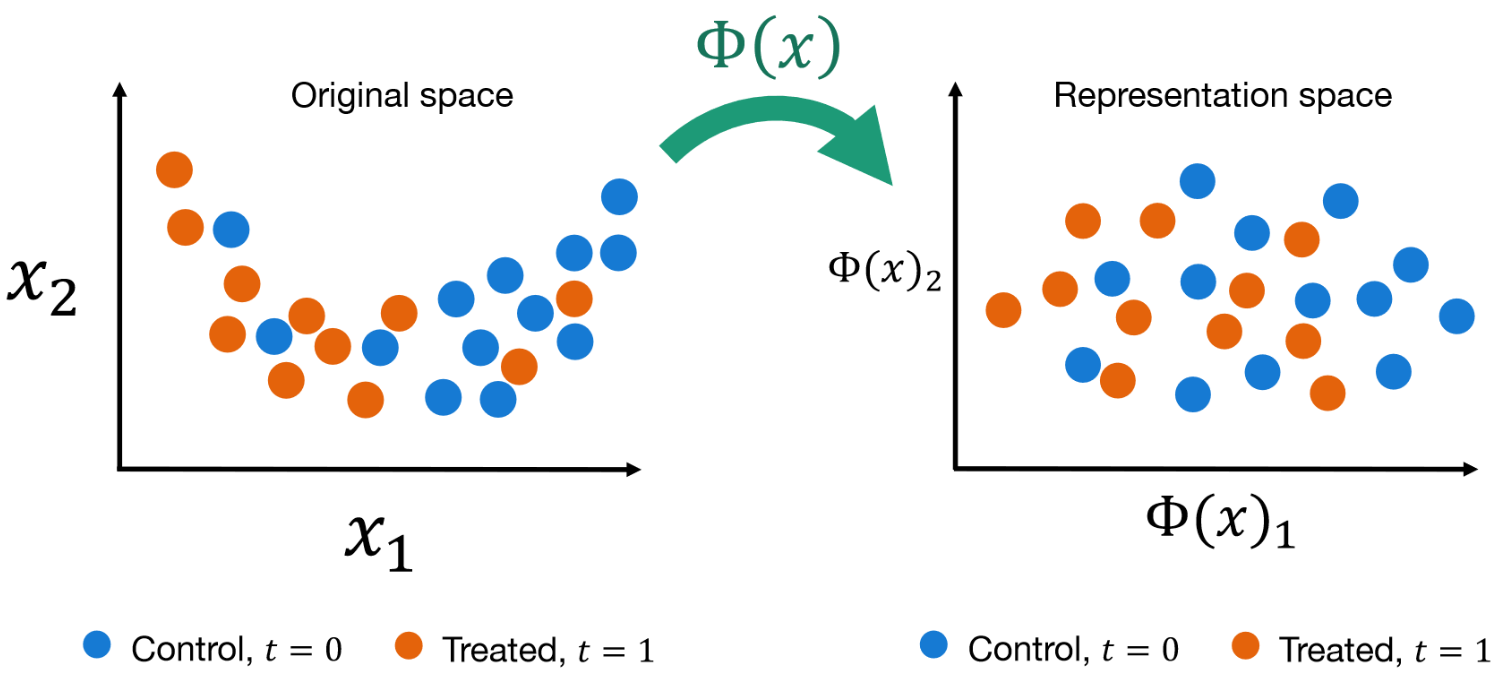}
    \caption{\textbf{Balancing through representation learning.} The promise of deep learning for causal inference is that a neural network encoding function $\Phi$ can transform the treated and control covariate distributions into a representation space such that they are indistinguishable. Used with permission from \citet{shenjohannson2018}.}
    \label{fig:balancing}
\end{figure}

\emph{The novel contribution of deep learning to causal estimation is the proposal that a neural network can learn a function $\Phi$ that produces representations of the covariates decorrelated from the treatment.} Fundamentally, the idea is that $\Phi$ can transform the treated and control covariate distributions into a representation space such that they are indistinguishable (Fig. \ref{fig:balancing}). To ensure that these representations are also still predictive of the outcome (multi-task learning), multiple loss functions are generally applied simultaneously to balance these objectives. This approach is applied in a majority of the algorithms presented in section \ref{section:mainbody}.

%% file: content/causalinference.tex
\section{Causal Identification and Estimation Strategies}
\label{section:CI}

\subsection{Identification of Causal Effects}

The papers described in this primer are primarily framed within the Potential Outcomes causal framework (Neyman-Rubin causal model) \citep{rubin1974estimating,imbens2015causal}. This framework is concerned with identifying the ``potential outcomes" of each unit $i$ in the sample, had it received treatment ($Y(1)$) or not received treatment ($Y(0)$). However, because each unit can only receive one treatment regime in reality (being treated or remaining untreated), it is not possible to observe both potential outcomes for each individual (often termed ``the fundamental problem of causal inference") \citep{holland1986statistics}. While we cannot thus identify individual treatment effects $\tau_i=Y_i(1)-Y_i(0)$ for each unit, causal inference frameworks allow us to probabilistically estimate average treatment effects ($ATE$) and average treatment effects conditional on select covariates ($CATE$) across samples of treated and control units. Within this literature, the motivation of many papers is to present algorithms that can both infer CATEs from observational data, but also predict them for out-of-sample units where treatment status is unknown. For readers unfamiliar with causal inference, a short introduction is glossed in Box \ref{box:causalintro} with a concrete example, used in the tutorials, in Box \ref{box:example}.

\begin{TCBBox}[label=box:causalintro]{Basic Introduction to Causal Inference}
 Correlation does not equal causation, and causal inference is concerned with the identification of causal relationships between random variables. Many causal questions we would like to ask about social data (``What is the causal effect of $T$ on $Y$ for units with characteristics $X$?") can be unpacked as counterfactual questions with the general format: ``What would have been the outcome $Y$ for a unit with $X$ characteristics, if $T$ had happened or not happened?".
 
Randomized control trials (RCTs, also known as A/B testing in data science and industry applications) are usually understood to be the ideal approach to answering this type of question: each unit with covariates or features $X$ is randomly assigned to the treatment or control groups and outcome $Y$ is subsequently measured. But in many scenarios it is prohibitively expensive or unethical (e.g., randomly assigning students to attend college or not) to collect experimental data. In these cases, we can statistically adjust observational data (e.g., survey data on college attendance) to approximate the experimental ideal. The methods described in this paper are designed to answer counterfactual questions with primarily non-experimental observational data. 

There are at least three different schools of causal inference that have been introduced in social statistics and econometrics \citep{rubin1974estimating,imbens2015causal}, epidemiology \citep{robins1986new,robins1987graphical,hernancausal2020}, and computer science \citep{goldszmidt1996qualitative,pearl2009causality}. The goal of these causal frameworks is to describe and correct for biases in data or study design that would prevent one from making a true causal claim. If these biases are correctable and the causal effect can be uniquely expressed in terms of the distribution of observed data, then we say that the causal effect is \textit{identifiable} \citep{kennedy2016semiparametric}.
 Only if a causal effect is identifiable, we can use statistical tools to correct for biases and \textit{estimate} the causal effect (e.g., inverse propensity score weighting, g-computation, deep learning). Other forms of bias, like sample selection bias, are possible to correct for, but are outside the focus of this article.

The algorithms presented in this paper focus on estimating causal effects primarily by correcting for \textit{confounding} bias. Loosely speaking, a confounding covariate/feature is one that is correlated with both the treatment and the outcome, misleadingly suggesting that the treatment has a causal effect on the outcome, or obscuring a true causal relationship between the treatment and outcome. Often times, the confounder is a cause of the treatment \textit{and} outcome. As an example of confounding bias, estimating the causal effect of attending college (treatment) on adult income (outcome) requires controlling for the fact that parental income may be a common cause of both college attendance and adult income. 
\end{TCBBox}

The ATE is defined as:
 
$$ATE =\mathbb{E}[Y_i(1)-Y_i(0)] = \mathbb{E}[\tau_i]$$
 
where $Y_i(1)$ and $Y_i(0)$ are the potential outcomes had the unit $i$ received or not received the treatment, respectively. The CATE is defined as,

$$CATE =\mathbb{E}[Y_i(1)-Y_i(0)|X_i=x] = \mathbb{E}[\tau_i|X_i=x] $$
 
where $X$ is the set of selected, observable covariates, and $x \in X$.

\begin{TCBBox}[label=box:example]{Applied Causal Inference Example: The Infant Health and Development Study}
To make this problem setting more concrete for readers unfamiliar with causal inference, consider simulations based on the 1985-1988 Infant Health and Development Study that are widely used as benchmarks within this literature. In this experiment, premature children were randomly assigned to intensive, high-quality childcare ($T$), and their cognitive test scores were measured later ($Y$). The authors also measured numerous other covariates $X$ including ``pregnancy complications, child's birth weight and gestation age, birth order, child's gender, household composition, day care arrangements, source of health care, quality of the home environment, parents' race and ethnicity, and maternal age, education, IQ, and employment" \citep{gross1993infant}. The $ATE$ would be the effect of intensive child care on cognitive scores across all children, while various $CATE$s might be formulated to better understand how the effects of child care vary for female children, children born to teenage mothers, or children with unemployed parents.   

\citet{Hill2011} turns this experimental data into an observational benchmark by re-simulating the outcome such that the covariates $X$ induce confounding bias between the treatment and outcome. While the simulations don't preserve the names of the covariates, we can imagine some confounding relationships that might be present in an observational study. For example, suppose that affluent ($X_1$) parents are more likely able to afford high-quality child care ($T$), but there is actually a weak association between childcare and premature babies' cognitive ability ($Y$). We also know affluent parents are more likely to engage in breastfeeding ($X_2$), which is positively associated with higher cognitive ability \citep{heck2006socioeconomic,kramer2008breastfeeding}. If we do not account for the correlation between income and childcare ($X_1 \rightarrow T$), or income and cognitive ability ($X_1 \rightarrow X_2 \rightarrow Y$), we may have bias in our $ATE$/$CATE$ estimates, or worse, erroneously interpret the correlation between childcare and cognitive ability as causal. This example is depicted in a causal graph below.

\begin{center}
\includegraphics[width=.7\textwidth]{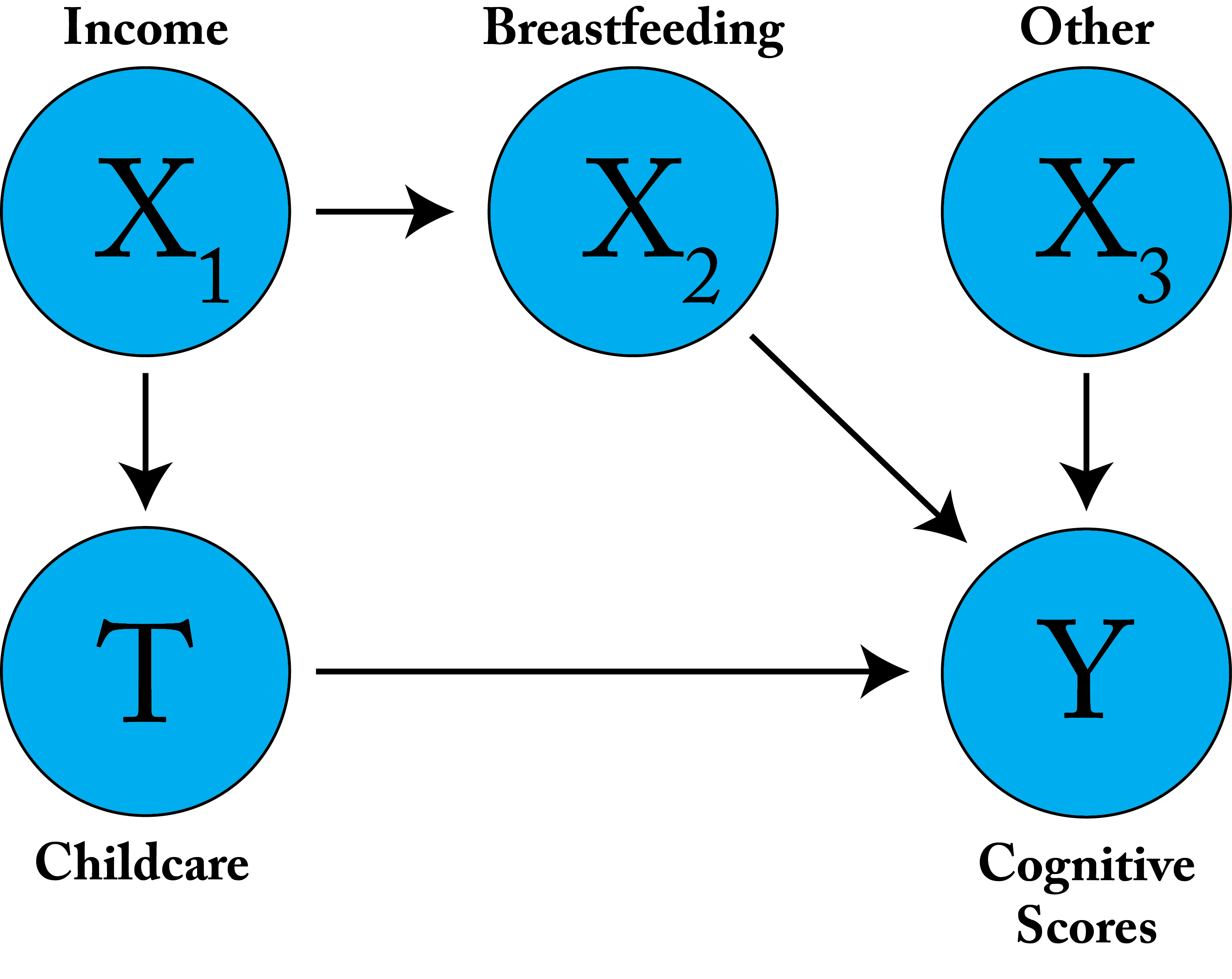}
\end{center}

The hypothetical confounding bias presented here can be adjusted for either through treatment modeling (e.g., inverse propensity score weighting, non-parametric, deep representation learning) to block the path $X_1 \rightarrow T$, outcome modeling (e.g., generalized linear models, deep regression) to block the path $X_1 \rightarrow X_2 \rightarrow Y$, or both (see Section \ref{sec:estimationstrats}). For coded examples using many of these approaches in Tensorflow and Pytorch on the IHDP benchmark, please see the tutorials.

\end{TCBBox}

Within the machine learning literature on causal inference treated here, the primary strategy for causal identification is \textbf{selection on observables}. A challenge to identifying causal effects is the presence of confounding relationships between covariates associated with both the treatment and the outcome. 

The key assumptions allowing the identification of causal effects in the presence of confounding is:

1. \textbf{Conditional Ignorability/Exchangability} The potential outcomes $Y(0)$, $Y(1)$ and the treatment $T$ are conditionally independent given $X$,
$$Y(0),Y(1)\perp \!\!\! \perp T|X $$
Conditional Ignorability specifies that there are no unmeasured confounders that affect both treatment and outcome outside of those in the observed covariates/features $X$. Additionally $X$ may contain predictors of the outcome (helping precision), but should not contain instrumental variables (hurting precision and potentially amplifying residual bias) or colliders within the conditioning set.\footnote{A variable is a collider if it is caused by two other variables. Controlling for colliding variables, or descendants of colliding variables, will induce a spurious correlation between the parents. In the case of adjusting for confounding, controlling for a collider variable can (re-)open a confounding path that would otherwise be closed, introducing additional bias.}

Other standard assumptions invoked to justify causal identification are:

2. \textbf{Consistency/Stable Unit Treatment Value Assumption (SUTVA)}. Consistency specifies that when a unit receives treatment, their observed outcome is exactly the corresponding potential outcome (and the same goes for the outcomes under the control condition). Moreover, the response of any unit does not vary with the treatment assignment to other units (i.e., no network or spillover effects), and the form/level of treatment is homogeneous and consistent across units (no multiple versions of the treatment). Note that this is an identification assumption, based on our understanding of the data generating process, and independent of the model chosen for estimation. More formally,
$$T=t \rightarrow Y=Y(T)$$

3. \textbf{Overlap}. For all $x \in X$ (i.e., any observed covariate value), all treatments $t\in \{0,1\}$ have a non-zero probability of being observed in the data, within the ``strata" defined by such covariates, 
$$1 > p(T=t|X=x)  > 0$$

4. An additional assumption sometimes invoked at the interface of identification and estimation using neural networks is:

\textbf{Invertability}
$$\Phi^{-1}(\Phi(X)) = X$$
In words, there must exist an inverse function of the representation function $\Phi$ encoded by a neural network that can reproduce $X$ from representation space. This is required for the Conditional Ignorability assumption to hold when using representation learning. From a practical perspective, it also means that the representation we created is rich enough to capture the causal relationships we are interested in.

For reference, we describe the full notation used within the review in Box \ref{Box:Notation}.
\begin{TCBBox}[label=Box:Notation]{Notation for Causal Inference and Estimation}
\
 We use uppercase to denote general quantities (e.g., random variables) and lowercase to denote specific quantities for individual units (e.g., observed variable values).
\\
\textbf{Causal identification}
\begin{small}
\begin{itemize}
\item Observed covariates/features: $X$
\item Potential outcomes: $Y(0)$ and $Y(1)$
\item Treatment: $T$
\item Unobservable Individual Treatment Effect: $\tau_i=Y_i(1)-Y_i(0)$
\item Average Treatment Effect: $ATE =\mathbb{E}[Y_i(1)-Y_i(0)] = \mathbb{E}[\tau_i]$
\item Conditional Average Treatment Effect: $CATE(x) =\mathbb{E}[Y_i(1)-Y_i(0)|X_i=x] = \mathbb{E}[\tau_i|X_i=x]$
\end{itemize}
\end{small}
\textbf{Deep learning estimation}
\begin{small}
\begin{itemize}
\item Predicted potential outcomes: $\hat{Y}(0)$ and $\hat{Y}(1)$
\item Outcome modeling functions: $\hat{Y}(T)=h(X,T)$ 
\item Propensity score function: $\pi(X,T)=P(T|X)$ (where $\pi(X,0)=1-\pi(X,1)$)
\item Representation functions: $\Phi(X)$ (producing representations $\phi$)
\item Loss functions: $\mathcal{L}(true,predicted)$
\item Loss abbreviations: $MSE$ (mean squared error), $BCE$ (binary cross-entropy),
CCE (categorical cross-entropy)
\item Loss hyperparameters: $\lambda,\alpha,\beta$
\item Estimated CATE*: $\hat{CATE_i} = \hat{\tau}_i = \hat{Y_i}(1)-\hat{Y_i}(0) = h(X,1)-h(X,0)$
\item Estimated ATE: $\hat{ATE}=\frac{1}{N}\sum_{i=1}^N\hat{\tau_i}$
\end{itemize}
\end{small}
Beyond the $ATE$ and $CATE$ there is an additional metric commonly used in the machine learning literature, first introduced by \citet{Hill2011} called the \textit{Precision in Estimated Heterogeneous Effects (PEHE)}. PEHE is the average error across the predicted $CATE$s.
\begin{itemize}
\small \item Precision in Estimated Heterogeneous Effects: $PEHE=\frac{1}{N}\sum_{i=1}^N(\tau_i-\hat{\tau_i})^2$
\end{itemize}
Beyond being a metric for simulations with known counterfactuals, the $PEHE$ has theoretical significance in the formulation of generalization bounds within this literature \citep{Shalit2017,Johansson2018,johannson2020,zhang2020learning}.

*Note that we use $\hat{\tau}$ to refer to the estimated CATE because truly individual treatment effects cannot be described only by the observed covariates $X$. 
\end{TCBBox}

\subsection{Estimation of Causal Effects}\label{sec:estimationstrats}

\begin{figure}[!p]
    \centering
    \includegraphics[width=\textwidth]{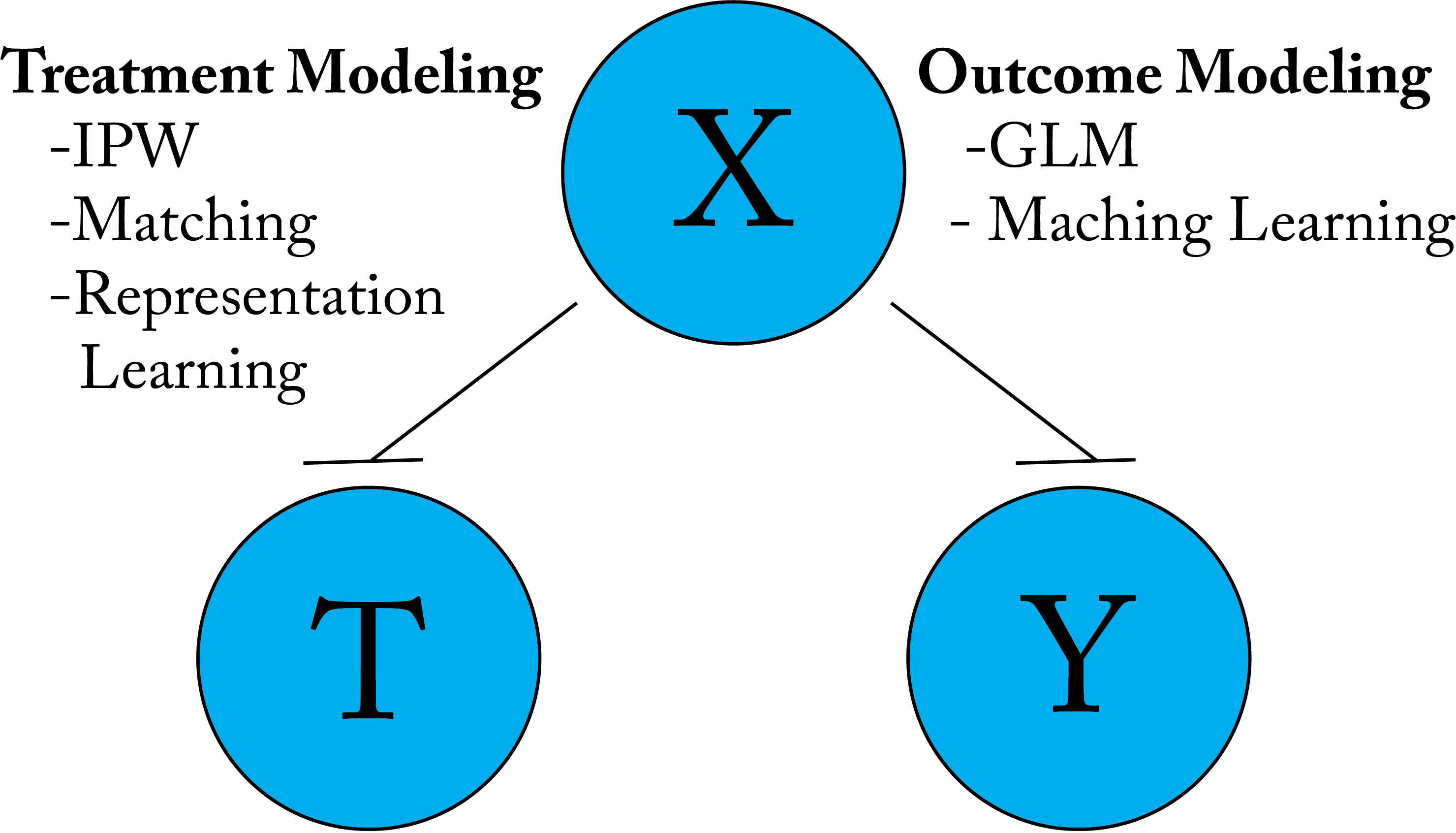}
    \caption{\textbf{Two fundamental approaches to deconfounding.} Blunted arrows indicate blocked causal paths. Treatment modeling approaches like inverse propensity weighting, balancing, and representation learning adjust for the association between the covariates $X$ and the treatment $T$. Outcome modeling approaches like generalized linear models or machine learning regressors adjust for the association between $X$ and the outcome $Y$.}
\label{fig:modelingstrats}
\end{figure}

Once a strategy for identifying causal effects from available data has been developed (arguably the harder and more important part of causal inference), statistical methods can be used to estimate causal effects by controlling for confounding bias, selection bias, and/or measurement error. There are two fundamental approaches to estimation: \textbf{treatment modeling} to control for correlations between the covariates $X$ and the treatment $T$, and \textbf{outcome modeling} to control for correlations between the treatment $X$ and the outcome $Y$ (Fig. \ref{fig:modelingstrats}). Below we briefly review three traditional techniques for removing confounding bias to motivate our systematization of deep learning models. First, we discuss outcome modeling through regression. Next, we consider treatment modeling through non-parametric matching. Finally, we discuss treatment modeling through inverse propensity score weighting (IPW) and introduce the concept of double robustness. 

\subsubsection{Outcome Modeling: Regression}

Assuming the treatment effect is constant across covariates/features or the probability of treatment is constant across all covariates/features (both improbable assumptions), the simplest consistent approach to estimating the $ATE$ is to regress the outcome on the treatment indicator and covariates using a linear model.\footnote{Another outcome modeling approach that could be used to estimate the outcome, not discussed here, is g-computation \citep{robins1986new,hernancausal2020}.} The ATE is then the coefficient of the treatment indicator. Without loss of generality, we call outcome models of this nature, linear or non-linear, $h$:

$$\hat{Y}_i(T)=h(X_i,T)$$

A slightly more sophisticated semi-parametric approach to \textbf{outcome modeling}, used widely in the application of machine learning to causal inference, is to use $h(X,T)$ to impute $\hat{Y}(1)$ and $\hat{Y}(0)$, and calculate the CATE for each unit as a plug-in estimator:

$$\widehat{CATE_i}=\hat{\tau_i}=\hat{Y_i(1)}-\hat{Y_i(0)}=h(X_i,1)-h(X_i,0)$$

and the ATE as:

$$\widehat{ATE}=\frac{1}{N}\sum_{i=1}^N{\hat{\tau_i}}$$

\subsubsection{Treatment Modeling: Non-Parametric Matching}
\label{subsection:doublyrobust}

A common treatment-modeling strategy is balancing the treated and control covariate distributions through matching. Matching requires the analyst to select a distance measure that captures the difference in observed covariate distributons between a treated and untreated unit \citep{austin2011introduction}. Units with treatment status $T$ can then be matched with one or more counterparts with treatment status $1-T$ using a variety of algorithms \citep{stuart2010matching}. In a one-to-one matching scenario where each treated unit has an otherwise identical untreated counterpart, the covariate distribution of treated and control units is indistinguishable. 

\subsubsection{Treatment Modeling: Inverse Propensity Score Weighting}
\label{subsection:treatmentmodeling}

Another common approach is \textbf{inverse propensity score weighting (IPW)}. In IPW, units are weighted on their inverse propensity to receive treatment. Without loss of generality, we call the propensity function $\pi$. The propensity score is calculated as the probability of receiving treatment conditional on covariates:

$$\pi(X,T)=P(T|X)$$

The simplest IPW estimator of the ATE is then:

 \begin{equation}\label{propscore}
    \widehat{ATE}=\frac{1}{N}\sum_{i=1}^N\left\{\frac{T_iY_i}{\hat{\pi}(X_i,1)}+\frac{(1-T_i)Y_i}{\hat{\pi}(X_i,0)}\right\}
\end{equation}

Note that only one of the two terms is active for any given unit. Furthermore, this presentation looks  different than how the IPW is generally presented because we use $\pi$ as a function with different outputs depending on the value of $T$ rather than a scalar (Box \ref{Box:Notation}).\footnote{To de-emphasize the contribution of units with extreme weights due to sparse data, sometimes a ``stabilized" IPW is used \citep{glynn2010introduction}.}

IPW weighting is attractive because if the propensity score $\pi$ is specified correctly, it is an unbiased estimator of the ATE. Moreover, the IPW is consistent if $\pi$ is estimated consistently \citep{rosenbaum1983central,glynn2010introduction}.

\subsubsection{Double Robustness}
Because different models make different assumptions, it is not uncommon to combine outcome modeling with propensity modeling or matching estimators to create \textbf{doubly-robust} estimators. For example, one of the most widely used doubly-robust estimators is the Augmented-IPW (AIPW) estimator. 


\begin{equation}
\hat{ATE} = \frac{1}{N}\sum_{i=1}^N{[\underbrace{\underbrace{(\frac{T}{\pi(\Phi(X),1)}-\frac{1-T}{\pi(\Phi(X),0)})}_{\text{Treatment Modeling}}\times\underbrace{[Y-h(\Phi(X),T)]}_{\text{Residual Confounding}}}_{\text{Adjustment}} +\underbrace{[h(\Phi(X),1)-h(\Phi(X),0)]}_{\text{Outcome Modeling}}}
\end{equation}

The first term is the difference in prediction from two outcome models, one for treated and one for 
control units, while the last terms is a ``corrected'' IPW estimator replacing the raw outcome by the residuals from the regression models. As expected, this estimator is unbiased if the IPW and regression estimators are consistently estimated. However, the model is attractive because it will be consistent if \textit{either} the propensity score $\pi(X,T)$ is correctly specified or the regression model $h$ is consistently specified \citep{glynn2010introduction}. The model also provide efficiency gains with respect to the use of each model separately, and especially with respect to weighting alone. 

Doubly robust estimation is especially important for causal estimation using machine learning. When using simple outcome plug-in estimators, bias is directly dependent on estimation error, which may be different for each potential outcome depending on the modeling strategy \citep{kennedy2020}. Machine learning estimation of the propensity score can also rely heavily on non-confounding predictors, giving rise to extreme weights \citep{schnitzer2016variable}. More generally, there are no asymptotic linearity guarantees for machine learning estimators which may converge at a slow rate, leading to misleading confidence intervals \citep{naimi2021,zivich2021machine}. For these reasons, plug-in machine learning estimation often has poor empirical performance when not using double robust estimators \citep{benkeser2017,kennedy2020,zivich2021machine}.

The growth of machine learning for causal
inference literature has thus been largely driven by the introduction of semi-parametric frameworks. Semi-parametric frameworks  address these issues by using machine learning only to estimate the nuissance parameters (i.e., potential outcomes and propensity score) of influence functions for causal parameters like the ATE and CATE \citep{chernozhukov2016double,kennedy2016semiparametric,van2011targeted}. In these approaches, the estimation of causal parameters is only-second order dependent on machine learning error, there is double-robustness against inconsistent estimation, and guarantees of fast convergence and asymptotically-valid confidence intervals even if the machine learning models converge slowly \citep{benkeser2017,kennedy2020,naimi2021,zivich2021machine}. We use the final algorithm introduced below, Dragonnet, as an opportunity to provide an intuitive introduction to semi-parametric theory and how it can be used for doubly robust estimation \citep{Shi2019}.

%% file: content/architectures.tex
\section{Three Different Approaches to Deep Causal Estimation}
\label{section:mainbody}

The architectures proposed in the deep learning literature for causal estimation build upon the core idea discussed above. First, we introduce ``S-Learners" and ``T-Learners" to show how neural networks can be used to estimate non-linearities in potential outcomes. Second, given the right objectives, a neural network can learn representations of the treated and control distributions that are deconfounded (Fig. \ref{fig:balancing}). This approach, which can be related theoretically to non-parametric matching, is illustrated by the foundational TARNet algorithm in section \ref{section:ipw} \citep{Shalit2017}. Finally, the machine learning for causal inference literature has been largely driven by the introduction of semi-parametric frameworks that allow predictive machine learning models to be plugged-in to doubly robust estimation equations \citep{van2011targeted,chernozhukov2016double,chernozhukov2021automatic}. In section \ref{section:ipw}, we introduce the concept of influence functions and the targeted maximum likelihood estimator to explain the Dragonnet algorithm. For clarity the algorithms presented here all share a familial resemblence to the TARNet algorithm. However, we note that there are many other approaches to using deep learning for causal inference (e.g., the generative models described in Appendix A.\ref{appendix:generative}).

%% file: content/outcomemodeling.tex
\subsection{Deep Outcome Modeling}
\label{section:outcomemodeling}

\emph{S-Learners and T-Learners (\textit{Tutorial 1 \href{https://colab.research.google.com/drive/1hjnyfJjFm0wWM3BcZMi0cpW0uBRd5c5f?usp=sharing}{\includegraphics[height=\fontcharht\font`\B,keepaspectratio]{content/colab_button.png} }})}

\begin{figure}
    \centering
    \includegraphics[width=\linewidth]{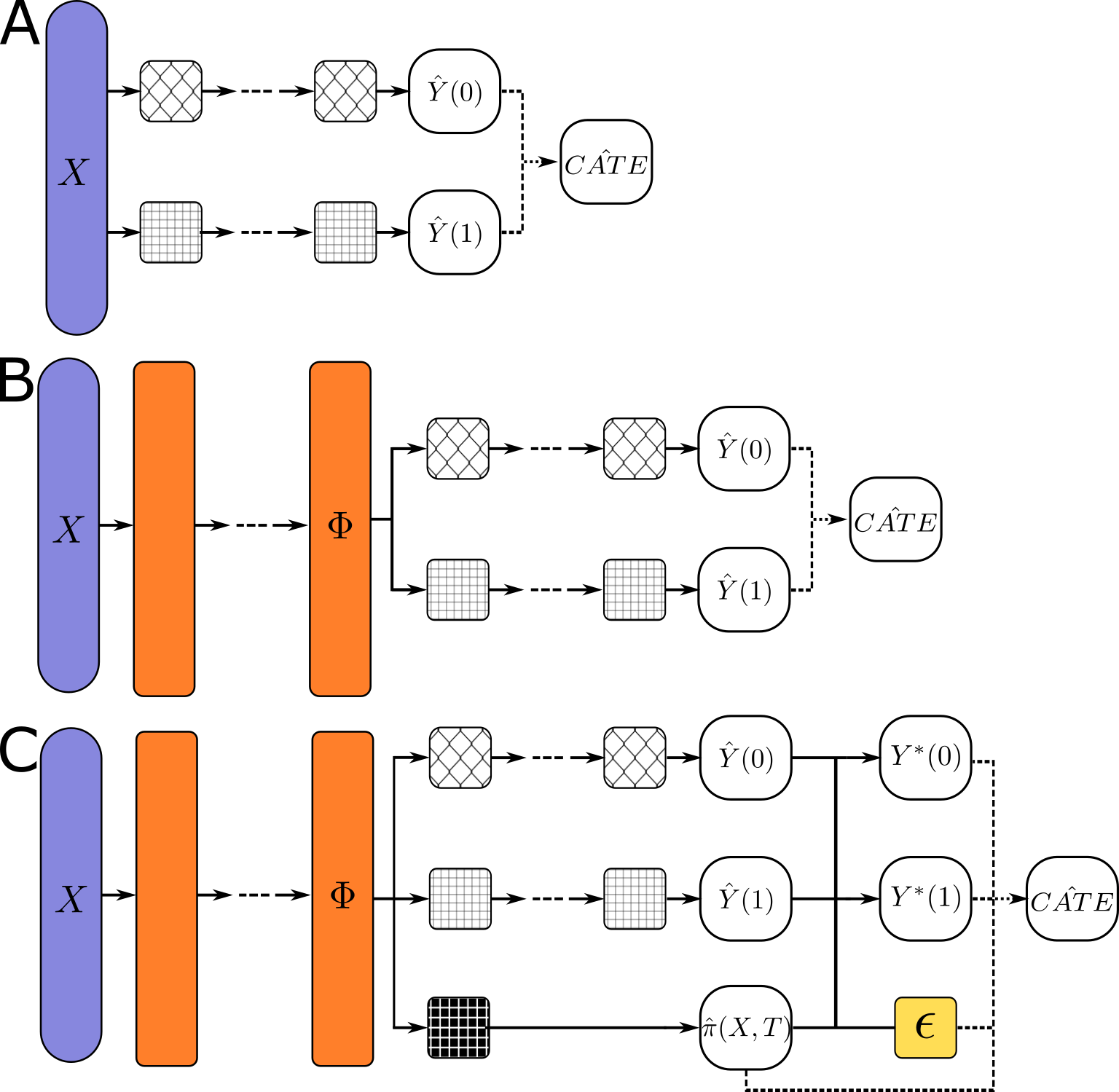}
    \caption{A. \textbf{T-learner.} In a T-learner, separate feed-forward networks are used to model each outcome (textured boxes). We denote the function encoded by these outcome modelers $h$. B. \textbf{TARNet.} TARNet extends the T-learner with shared representation layers (orange). The motivation behind TARNet (and further elaborations of this model) is that the multi-task objective of accurately predicting both the treated and control potential outcomes forces the representation layers to learn a balancing function $\Phi$ such that the $\Phi(X|T=0)$ and $\Phi(X|T=1)$ are overlapping distributions in representation space. For a code implementation, see Box \ref{box:tarnetcode}. \textbf{C. Dragonnet} Dragonnet also adds a propensity score head to TARNet (black textured box) and a free ``nudge" parameter $\epsilon$. In an adaptation of Targeted Maximum Likelihood Estimation, $\hat{\pi}$ and $\epsilon$ are used to re-weight the outcomes to provide lower biased estimates of the $ATE$.}
    \label{fig:tlearner}
\end{figure}

Because at most one potential outcome is  unobserved, it is not possible to apply supervised models to directly learn treatment effects. Across econometrics, biostatistics, and machine learning, a common approach to this challenge has been to instead use machine learning to model each potential outcome separately and use plug-in estimators for treatment effects \citep{chernozhukov2016double,van2011targeted,wager2018}. As with linear models, a single neural model can be trained to learn both potential outcomes (``S[ingle]-learner") (Fig. \ref{fig:perceptron}B), or two independent models can be trained to learn each potential outcome (a ``T-learner") \citep{johannson2020} (Fig. \ref{fig:tlearner}A). In both cases, the neural network estimators would be feed-forward networks tasked with minimizing the MSE in the prediction of observed outcomes. In a slight abuse of notation, the joint loss function for a T-learner can be written as:
\begin{equation}
\mathcal{L}(Y,h(X,T))=MSE(T_i(Y_i,h_1(X_i,1))+(1-T_i)(Y_i,h_0(X_i,0))
\end{equation}
where $h_1$ and $h_0$ represent separate networks for each potential outcome. 

After training, inputting the same unit into both networks of a T-learner will produce predictions for both potential outcomes: $\hat{Y}(T)$ and $\hat{Y}(1-T)$. We can plug-in these predictions to estimate the $CATE$ for each unit, 
$$\hat{\tau_i}=(1-2T_i)(\hat{Y_i}(1-T_i)-\hat{Y_i}(T_i))$$
where the first term is a switch to make sure the treated potential outcome comes first. The average treatment effect as,
$$\widehat{ATE}=\frac{1}{N}\sum_{i=1}^N\hat{\tau_i}$$

Nearly all of the models described below combine this plug-in outcome modeling approach with other forms of treatment adjustment.

%% file: content/balancing.tex
\subsection{Balancing through Representation Learning}
\label{section:balancing}

\emph{TARNet (\textit{Tutorial 1 \href{https://colab.research.google.com/drive/1hjnyfJjFm0wWM3BcZMi0cpW0uBRd5c5f?usp=sharing}{\includegraphics[height=\fontcharht\font`\B,keepaspectratio]{content/colab_button.png} }})}
\label{section:TARNet}

Balancing is a treatment adjustment strategy that aims to deconfound the treatment from outcome by forcing the treated and control covariate distributions closer together \citep{Johansson2016}. \textit{The novel contribution of deep learning to the selection on observables literature is the proposal that a neural network can transform the covariates into a representation space $\Phi$ such that the treated and control covariate distributions are indistinguishable (Fig. \ref{fig:balancing}).}

To encourage a neural network to learn balanced representations, the seminal paper in this literature, \citet{Shalit2017}, proposes a simple two-headed neural network called Treatment Agnostic Regression Network (TARNet) that extends the outcome modeling T-learner with shared representation layers (Fig. \ref{fig:tlearner}B). Each head models a separate potential outcome: one head learns the function $\hat{Y}(1)=h_1(\Phi(X),1)$, and the other head learns the function $\hat{Y}(0)=h_0(\Phi(X),0)$. During training, only one head will receive error gradients at a time (the one predicting the observed outcome). However, both heads backpropagate their gradients to shared representation layers that learn $\Phi(X)$. The idea is that these representation layers must learn to balance the data because they are tasked with predicting both outcomes. The authors of this algorithm have subsequently extended TARNet with additional losses in an algorithm called CFRNET that explicitly encourage balancing by minimizing a statistical distance between the two covariate distributions in representation space (see Appendix A.\ref{appendix:cfrnet} for details) \citep{Johansson2018,johannson2020}. 

The complete objective for the network is to fit the parameters of $h$ and $\Phi$ for all $n$ units in the training sample such that,
\begin{equation}
\argmin_{h,\Phi}\frac{1}{N}\sum_{i=1}^N (Y_i -(T_i(Y_i,\underbrace{h_1(\Phi(X_i),1)}_{\hat{Y_i}(1)}) + (1-T_i)(Y_i,\underbrace{h_0(\Phi(X_i),0)}_{\hat{Y_i}(0)})^2 + \lambda\underbrace{\mathcal{R}(h)}_{L_2} 
\end{equation}

 or more compactly,
\label{eq:tarnetloss}
\begin{equation}
\argmin_{h,\Phi} MSE(Y_i,\underbrace{h(\Phi(X_i),T_i)}_{\hat{Y_i}(T_i)}) + \lambda\underbrace{\mathcal{R}(h)}_{L_2} 
\end{equation}
where $\mathcal{R}(h)$ is a model complexity term (e.g., for $L_2$ regularization) and $\lambda$ is a hyperparameter chosen through model selection. For coded versions of TARNet in Tensorflow and Pytorch, see Box \ref{box:tarnetcode}.  

\begin{TCBBox}[label=box:tarnetcode]{TARNet in Code}
Below we show simple implementations of TARNet in Python Tensorflow 2 and Pytorch. For more explanation on this implementation and to run this code on the IHDP data, see the tutorials.

\textbf{Tensorflow 2 Functional API (Keras)}
\begin{python}
def make_tarnet(input_dim):
    #The argument is the number of X covariates.
    x = Input(shape=(input_dim,), name='input')
    
    #In TF fxnl API, stack layers by feeding output of prev layer to next
    #Make 2 representation layers
    #units is the output dim of layer
    #elu is \"exponentiated linear unit" activation fxn
    phi = Dense(units=200, activation='elu')(x)
    phi = Dense(units=200, activation='elu')(phi)

    #Begin separate outcome modeling heads
    y0_hidden = Dense(units=100, activation='elu')(phi)
    y1_hidden = Dense(units=100, activation='elu')(phi)

    # Add second layers
    y0_hidden = Dense(units=100, activation='elu')(y0_hidden)
    y1_hidden = Dense(units=100, activation='elu')(y1_hidden)

    # Output predictions
    y0_pred = Dense(units=1, activation=None)(y0_hidden)
    y1_pred = Dense(units=1, activation=None)(y1_hidden)
    
    #Bundle outputs
    concat_pred = Concatenate(1)([y0_pred, y1_pred])
    
    #instantiate model
    model = Model(inputs=x, outputs=concat_pred)
    return model
\end{python}
\textbf{Pytorch}
\begin{python}
class TARNet(nn.Module):
    def __init__(self,input_dim):
        super(TARNet,self).__init__()
        
        self.phi = nn.Sequential(
            #both input and output dims are specified in torch
            nn.Linear(input_dim, 200),
            nn.ELU(), #activations are discrete from layers
            nn.Linear(200,200),
            nn.ELU())
            
        self.y0_hidden = nn.Sequential(
            nn.Linear(200, 100),
            nn.ELU(),
            nn.Linear(100,100),
            nn.ELU())
            
        self.y1_hidden = nn.Sequential(
            nn.Linear(200, 100),
            nn.ELU(),
            nn.Linear(100,100),
            nn.ELU())
            
        self.y0_pred =nn.Linear(100,1)
        self.y1_pred = nn.Linear(100,1)
        
    #the flow of data/gradients in torch is declared in a forward fxn    
    def forward(self,X):
        rep = self.phi(X)
        y0_rep=self.y0_hidden(rep)
        y0_hat=rep=self.y0_pred(y0_rep)
        
        y1_rep=rep=self.y1_hidden(rep)
        y1_hat=rep=self.y1_pred(y1_rep)
        
        return y0_hat, y1_hat
\end{python}
\end{TCBBox}

%% file: content/propensity.tex
\subsection{Double Robustness with Inverse Propensity Score Weighting}
\label{section:ipw}

Rather than applying losses directly to the representation function, IPW methods estimate propensity scores from representations using the function $\pi(\Phi(X),T)=P(T|\Phi(X))$. As in traditional IPW estimators, these methods exploit the sufficiency of correctly-specified propensity scores to reweight the plugged-in outcome predictions and provide unbiased estimates of the ATE \citep{rosenbaum1983central}. Because these models combine outcome modeling with IPW, they retain the attractive statistical properties of doubly robust estimators discussed in section \ref{subsection:doublyrobust} \citep{Atan2018}. In this section we focus on \citet{Shi2019}'s Dragonnet model, which adapts semi-parametric estimation theory for batch-wise neural network training in a procedure they call ``Targeted Regularization" (TarReg) \citep{kennedy2016semiparametric}. Given the increasing importance of semi-parametric theory and ``double machine learning" across the causal estimation literature, we include a brief introduction to semi-parametric theory and targeted maximum likelihood estimation (TMLE) before diving into the details of the Dragonnet algorithm \cite{van2011targeted,chernozhukov2016double}.

\emph{Dragonnet (\textit{Tutorial 3 \href{https://colab.research.google.com/drive/1XzyOINgdSr78_KT7HJRJn07AwkHh-fG8?usp=sharing}{\includegraphics[height=\fontcharht\font`\B,keepaspectratio]{content/colab_button.png} }} \textit{/} \textit{Tutorial 4 \href{https://colab.research.google.com/drive/1NHYTbvGq-cWyy-mm0TrBH2rAmMqtcgPJ?usp=sharing}{\includegraphics[height=\fontcharht\font`\B,keepaspectratio]{content/colab_button.png} }})}

    A trivial extension to TARNet is to add a third head to predict the propensity score. This third head could use multiple neural network layers or just a single neuron, as proposed in Dragonnet (Fig. \ref{fig:tlearner}C) \citep{Shi2019}. Dragonnet uses this additional head to develop a training procedure called ``Targeted Regularization"  for semi-parametric causal estimation, inspired by ``Targeted Maxmimum Likelihood Estimation" (TMLE)\citep{van2011targeted}.

With three heads, the basic loss function for this network looks like:
\begin{equation}
\underset{\Phi,\pi,h}{\arg \min}\
\underbrace{MSE(Y_i,h(\Phi(X_i),T_i)}_{\text{Outcome Loss}} + \alpha \underbrace{\text{BCE}(T_i,\pi(\Phi(X_i),T_i))}_{\pi\text{ Loss}} +  \lambda\underbrace{\mathcal{R}(h)}_{L_2} 
\end{equation}
with $\alpha$ being a hyperparameter to balance the two objectives. The mean squared error and binary cross-entropy are standard objective functions in machine learning for regression and binary classification, respectively. Note that the first term is simply an expansion of the first term in equation \ref{eq:tarnetloss} 

Below, we explore how the authors add a second loss on top of this one to allow for semi-parametric estimation. 

\subsubsection{Semi-parametric Theory of Causal Inference}

In recent years, semi-parametric theory has emerged as a dominant theoretical framework for applying machine learning algorithms, including neural networks, to causal estimation \citep{chernozhukov2016double,chernozhukov2021automatic,chernozhukov2022riesznet,farrell2021deep,kennedy2016semiparametric,nie2021quasi,van2011targeted,wager2018}. The great appeal of these frameworks is that they allow for machine learning algorithms to be plugged-in for non-linear estimates of outcomes and propensity score, while still providing attractive statistical guarantees (e.g., consistency, efficiency, asymptotically-valid confidence intervals).

At a very intuitive level, semi-parametric causal estimation is focused on estimating a target parameter of a distribution $P$ (the $ATE$) of treatment effects $T(P)$ \citep{fisher2021visually}. While we do not know the true distribution of treatment effects because we lack counterfactuals, we do know some parameters of this distribution (e.g., the treatment assignment mechanism). We can encode these constraints in the form of a likelihood that parametrically defines a set of possible approximate distributions  $\mathcal{P}$ from our existing data $P$. Within this set there is a sample-inferred distribution $\tilde{P}\in\mathcal{P}$, that can be used to estimate $T(P)$ using $T(\tilde{P})$.

Regardless of $\tilde{P}$ chosen, $\tilde{P}\neq P \rightarrow T(\tilde{P})\neq T(P)$. We do not know how to pick $\tilde{P}$ with finite data to get the best estimate $T(\tilde{P})$. We can maximize a likelihood function to pick $\tilde{P}$, but there may be ``nuisance" parameters in the likelihood that are not the target and we do not care about estimating accurately. Maximum likelihood optimization may provide lower-biased estimates of these nuissance terms at the cost of better estimates of $T(P)$. 

To sharpen the likelihood's focus on $T(P)$, we define a ``nudge" parameter $\epsilon$ that moves $\tilde{P}$ closer to $P$ (thus moving $T(\tilde{P})$ closer to $T(P)$). An influence curve of $T(P)$ tells us how changes in $\epsilon$ will induce changes in $T(P+\epsilon(\tilde{P}-P))$. We'll use this influence curve to fit $\epsilon$ to get a better approximation of $T(P)$ within the likelihood framework. In particular, there is a specific \textbf{efficient influence curve (EIC)} that provides us with the lowest variance estimates of $T(P)$. In causal estimation, solving the EIC for the ATE yields estimates that are asymptotically unbiased, efficient, and have confidence intervals with (asymptotically) correct coverage.

The EIC for the ATE is,

\begin{equation}
EIC_{ATE} = \frac{1}{N}\sum_{i=1}^N{[\underbrace{\underbrace{(\frac{T_i}{\pi(X_i,1)}-\frac{1-T_i}{\pi(X_i,0)})}_{\text{Treatment Modeling}}\times\underbrace{(Y_i-h(X_i,T))}_{\text{Residual Confounding}}}_{\text{Adjustment}}] +\underbrace{[h(X_i,1)-h(X_i,0)]}_{\text{Outcome Modeling}}}]-ATE
\end{equation}
Setting $EIC_{ATE}$ to it's mean of 0,
\begin{equation}
ATE = \frac{1}{N}\sum_{i=1}^N{[\underbrace{\underbrace{(\frac{T_i}{\pi(X_i,1)}-\frac{1-T_i}{\pi(X_i,0)})}_{\text{Treatment Modeling}}\times\underbrace{(Y_i-h(X_i,T))}_{\text{Residual Confounding}}}_{\text{Adjustment}}] +\underbrace{[h(X_i,1)-h(X_i,0)]}_{\text{Outcome Modeling}}}]
\label{eq:eic}
\end{equation}
The underbraces illustrate how $EIC_{ATE}$ resembles a doubly robust estimator. When the EIC is minimized (set to 0) as in equation \ref{eq:eic}, the $ATE$ is equal to the outcome modeling estimate plus a treatment modeling estimate proportional to the residual error. 

\subsubsection{From TMLE to Targeted Regularization}

Targeted Regularization (TarReg) is closely modeled after ``Targeted Maxmimum Likelihood Estimation" (TMLE) \citep{van2011targeted}. 
TMLE is an iterative procedure where a nuissance parameter $\epsilon$ is used to nudge the outcome models towards sharper estimates of the ATE when minimizing the EIC as in Equation \ref{eq:eic}.\footnote{For a deeper dive on targeted learning, we recommend \citep{chambaz2020}.}

\begin{enumerate}
\singlespacing
\item Fit $h$ by predicting outcomes (e.g., using TARNet) and minimizing $MSE(Y,h(\Phi(X),T))$
\item Fit $\pi$ by predicting treatment (e.g., using logistic regression) and $BCE(T,\pi(\Phi(X),T))$
\item Plug-in $h$ and $\pi$ functions to fit $\epsilon$ and estimate $h^{*}(X,T)$ where,
$$\underbrace{h^{*}(X_i,T_i)}_{Y^*}=\underbrace{h(\Phi(X_i),\Phi(T_i))}_{\hat{Y}}+\underbrace{\left(\frac{T_i}{\pi(\Phi(X_i),1)}-\frac{1-T_i}{\pi(\Phi(X_i),0)}\right)}_{\text{Treatment Modeling Adjustment}}\times \underbrace{\epsilon}_{\text{``nudge"}}$$
by minimizing $MSE(Y,h^{*}(\Phi(X),T))$. This is equivalent to minimizing the ``Adjustment" part in equation \ref{eq:eic}.
\item Plug-in $h^*(X,T)$ to estimate $\hat{ATE}$:
$$\widehat{ATE}_{TMLE}=\frac{1}{N}\sum_{i=1}^N{ \underbrace{h^*(X_i,1)}_{Y_i^*(1)}-\underbrace{ h^*(X_i,0)}_{Y_i^*(0)}}$$
\doublespacing
\end{enumerate}

Targeted Regularization takes TMLE and adapts it for a neural network loss function. The main difference is that steps 1 and 2 above are done concurrently by Dragonnet, and that the loss functions for the first three steps are combined into a single loss applied to the whole network at the end of each batch. It requires adding a single free parameter to the Dragonnet network for $\epsilon$.

 At a very intuitive level, Targeted Regularization is appealing because it introduces a loss function to TARNet that explicitly encourages the network to learn the mean of the treatment effect distribution, and not just the outcome distribution. The Targeted Regularization procedure proceeds as follows:

In each epoch:
\begin{enumerate}
\singlespacing
    \item \begin{enumerate}
    \item Use Dragonnet to predict $h(\Phi(X),T)$ and $\pi(\Phi(X),T)$.
    \item Calculate the standard ML loss for the network using a hyperparameter $\alpha$:
    $$\underset{\Phi,\pi,h}{\arg \min}\ \underbrace{MSE(Y_i,h(\Phi(X_i),T_i))}_{\text{Outcome Loss}} + \alpha \underbrace{\text{BCE}(T_i,\pi(\Phi(X_i),T_i))}_{\pi\text{ Loss}} +  \lambda\underbrace{\mathcal{R}(h)}_{L_2}$$
    \end{enumerate}
    \item \begin{enumerate}
    \item Compute $h^{*}(\Phi(X_i),T_i)$ as above,
    $$\underbrace{h^{*}(\Phi(X_i),T_i)}_{Y^*}=\underbrace{h(\Phi(X_i),T_i)}_{\hat{Y_i}}+\underbrace{(\frac{T_i}{\pi(\Phi(X_i),1)}-\frac{1-T_i}{\pi(\Phi(X_i),0)})}_{\text{Treatment Modeling Adjustment}}\times \underbrace{\epsilon}_{\text{``nudge"}}$$
    \item Calculate the targeted regularization loss: $MSE(Y,h^*(\Phi(X),T))$
    \end{enumerate}
  \item Combine and minimize the losses from 1 and 2 using a hyperparameter $\beta$,
    $$\underset{\Phi,h,\epsilon}{\arg \min}= \underbrace{MSE(Y,h(\Phi(X),T))}_{\text{Outcome Loss}} + \alpha \cdot \underbrace{\text{BCE}(T,\pi(\Phi(X),T))}_{\pi\text{ Loss}}+  \lambda\underbrace{\mathcal{R}(h)}_{L_2}+\beta\cdot \underbrace{MSE(Y,h^*(\Phi(X),T))}_{\text{Targeted Regularization Loss}}$$
\doublespacing
\end{enumerate}

Step 3 of Targeted Regularization is exactly equivalent to minimizing the EIC up to a constant $\beta$. 

At the end of training, we can thus estimate the targeted regularization estimate of  the ATE $\hat{ATE_{TR}}$ as in TMLE:
$$\hat{ATE_{TR}}=\frac{1}{N}\sum_{i=1}^N{ \underbrace{h^*(\Phi(X_i),1)}_{Y^*_i(1)}-\underbrace{ h^*(\Phi(X_i),0)}_{Y^*_i(0)}}$$

Compared to S-learners, T-learners, and TARNet, the Dragonnet algorithm is particularly attractive because of the statistical guarantees afforded by its semiparametric framework. It is doubly robust, unbiased,  converges at a rate of $\frac{1}{\sqrt{n}}$, and the sampling distribution is asymptotically normal. Below we describe how to create assymptotically-valid confidence intervals for this estimator.

%% file: content/interpretation.tex
\section{Confidence and Interpretation}
\label{section:interpretation}

In this section, we move from theory to practice, and  treat best practices for building confidence intervals and interpreting heterogeneous treatment effects. Both of these topics are active areas of development, not only within the causal inference literature, but across machine learning research. Here we specifically focus on recommendations that can be easily implemented by analysts.

\subsection{Assessing Confidence}
\label{subsection:confidence}
(Tutorial 4 \href{https://colab.research.google.com/drive/1NHYTbvGq-cWyy-mm0TrBH2rAmMqtcgPJ?usp=sharing}{\includegraphics[height=\fontcharht\font`\B,keepaspectratio]{content/colab_button.png} })

In this paper, we feature Dragonnet over other approaches because of its attractive statistical properties. Because the Targeted Regularization procedure in Dragonnet is essentially a variant of TMLE, an asymptotically valid standard error can be calculated as the sample corrected variance of the efficient influence curve $\sigma_{\hat{ATE}}$, where
\begin{equation}
\sigma_{\hat{ATE_{TR}}}=\sqrt{\frac{Var(EIC_{\hat{ATE_{TR}}})}{N}}
\end{equation}
and,

\begin{equation}
Var(EIC_{\hat{ATE_{TR}}}) = Var[(\frac{T}{\pi(X,1)}-\frac{1-T}{\pi(X,0)})(Y-h^*(X,T))+(h^*(X,1)-h^*(X,0))-\hat{ATE_{TR}}]
\end{equation}
(\citep{van2011targeted}, pp. 96)

In Tutorial 5, we show how $\sigma_{\hat{ATE}}$ can be used to calculate a Wald confidence interval for Dragonnet. While not featured in this review, asymptotically valid conference intervals can also be calculated using RieszNet, a variant of Dragonnet introduced in \citet{chernozhukov2022riesznet} that connects neural network estimation to the automatically debiased machine learning literature currently popular in causal econometrics \citep{chernozhukov2016double,chernozhukov2021automatic}.

\subsection{Interpretation}
\label{subsection:interpretation}
(Tutorial 4 \href{https://colab.research.google.com/drive/1NHYTbvGq-cWyy-mm0TrBH2rAmMqtcgPJ?usp=sharing}{\includegraphics[height=\fontcharht\font`\B,keepaspectratio]{content/colab_button.png} })

A lack of interpretability has been a barrier to the adoption of machine learning methods like neural networks and random forests in social science settings. However, the literature on post-hoc interpretability techniques has matured considerably over the past five years, and several techniques for identifying important features/covariates such as permutation importance, LIME scores, SHAP scores, Individual Conditional Expectation plots etc... are in widespread usage today \citep{altmann2010,goldstein2017,lundberg2017,ribeiro2016should}. For a broad and accessible treatment on interpreting machine learning models, see \citet{molnar2022interpretable}.  

Building on criteria used to evaluate other explainable AI methods, \citet{crabbe2022benchmarking} note four desirable properties of a feature importance technique for the interpretation of deep causal estimators: sensitivity, completeness, linearity, and implementation invariance \citep{sundararajan2017axiomatic}.  A method that is 'sensitive' can distinguish between features that are simply predictive of the outcome, and those that actually influence CATE heterogeneity. A method that is 'complete' identifies all features that, together, explain all effect heterogeneity compared to a baseline. A 'linear' method is one where the feature importance scores additively describe the prediction. Lastly, the approach should be agnostic to both the model architecture (e.g., TARNet, Dragonnet) and different architectural hyperparameterizations (i.e., invariant to implementation). Of the feature importance methods surveyed, they identify two that manifest all four of these qualities: SHAP scores, and integrated gradients.

SHAP (SHapley Additive exPlanations) scores have emerged as one of the most popular methods for evaluating machine learning models in recent years \citep{lundberg2017}. SHAP is what is called a  “local” interpretability method: it provides feature importance estimates for each individual datum. Theoretically, SHAP frames feature importance estimation as a cooperative (game-theoretic) game between covariates to predict a specific outcome. Under the hood, the algorithm exhaustively compares all possible “coalitions” of covariates and their ability to predict the outcome (win the game). Predictions from this powerset of coalitions are used to calculate the additive marginal contributions of each feature in prediction using Shapley values. The disadvantage of SHAP is that, even with computational tricks, calculating scores for every unit can become computationally intractable in high dimensional datasets. SHAP scores are interpreted in comparison to a causal baseline of the ATE.

Because of the computational expense of SHAP scores, \citet{crabbe2022benchmarking} also recommend another local-interpretability method called “Integrated Gradients” \citep{sundararajan2017axiomatic}. Intuitively, this algorithm draws a straight-line, linear path in feature space between the target input (individual unit) and a baseline (i.e., a hypothetical unit who is exactly average on all covariates). A feature importance score can then be constructed by calculating the gradient in prediction error along this path with respect to the feature of interest. Note that SHAP scores can also be understood theoretically within the path framework. From this perspective, coalitions are paths in which each feature is turned on sequentially, and the SHAP score is the expectation across these paths. This interpretation leads to a gradient-based algorithm for calculating SHAP scores specifically for neural networks, which is also in the SHAP package. In practice, we recommend that analysts experiment with both integrated gradients and SHAP scores.

\subsection{What's in the tutorials?}
\label{subsection:tutorials}

To move from theory to empirics, the \href{https://github.com/kochbj/Deep-Learning-for-Causal-Inference/tree/main}{online tutorials} show how to implement many of the ideas presented throughout this primer. The tutorials are hosted in notebooks in the Google Colaboratory environment. When users open a Colab notebook, Google immediately provides a free virtual machine with standard Python machine learning packages available. This means that readers need not install anything on their own computers to experiment with these models. The tutorials are written in the Python programming language and provide examples in both Tensorflow2 and Pytorch, the two most popular deep learning frameworks. We note that both Tensorflow2 and Pytorch have implementations in R. However, we strongly recommend that readers interested in getting into deep learning work in Python, which has a much richer ecosystem of third-party packages for machine learning. 

Currently there are five tutorials:

\begin{itemize}
    \item \href{https://colab.research.google.com/drive/1hjnyfJjFm0wWM3BcZMi0cpW0uBRd5c5f?usp=sharing}{\includegraphics[height=\fontcharht\font`\B,keepaspectratio]{content/colab_button.png}} Tutorial 1 introduces S-learners, and T-learners before TARNet as a way to get familiar with building custom Tensorflow models.
    \item \href{https://colab.research.google.com/drive/1MBUkQO7rh89JrlAV0p1aLNoF9CIB3rZZ?usp=sharing}{\includegraphics[height=\fontcharht\font`\B,keepaspectratio]{content/colab_button.png}} Tutorial 2 focuses on causal inference metrics and hyperparameter optimization. Because we do not observe counterfactual outcomes, it's not obvious how to optimize supervised learning models for causal inference. This tutorial introduces some metrics for evaluating model performance. In the first part, you learn how to assess performance on these metrics in Tensorboard. In the second part, we hack \href{https://keras-team.github.io/keras-tuner/}{Keras Tuner} to do hyperparameter optimization for TARNet, and discuss considerations for training models as estimators rather than predictors.
    \item \href{https://colab.research.google.com/drive/1XzyOINgdSr78_KT7HJRJn07AwkHh-fG8?usp=sharing}{\includegraphics[height=\fontcharht\font`\B,keepaspectratio]{content/colab_button.png}} Tutorial 3 highlights the semi-parametric extension to TARNet featured in \citet{Shi2019}. We add treatment modeling to our TARNet model, and build an augmented inverse propensity score estimator. We then briefly describe the algorithm for Targeted Maximum Likelihood Estimation to introduce and build a Dragonnet with Shi et al.'s Targeted Regularization.  
    \item \href{https://colab.research.google.com/drive/1NHYTbvGq-cWyy-mm0TrBH2rAmMqtcgPJ?usp=sharing}{\includegraphics[height=\fontcharht\font`\B,keepaspectratio]{content/colab_button.png}} Tutorial 4 reimplements Dragonnet in Pytorch and shows how to calculate asymptotically-valid confidence intervals for the average treatment effect. We also interpret the features contributing to different heterogeneous CATEs using Integrated Gradients and SHAP scores. This tutorial is a good tutorial if you also just want to learn how to interpret SHAP scores, independent of the context of causal inference.
    \item \href{https://colab.research.google.com/drive/1phpoNEdvDgipCZCnpfvPC5anuIuodKB6?usp=sharing}{\includegraphics[height=\fontcharht\font`\B,keepaspectratio]{content/colab_button.png}} 
    Tutorial 5 features the Counterfactual Regression Network (CFRNet) and propensity-weighted CFRNet in \citet{Shalit2017,Johansson2018,johannson2020} (Appendix A.\ref{appendix:cfrnet}). This approach relies on integral probability metrics to bound the counterfactual prediction loss and force the treated and control distributions closer together. The weighted variant adds adaptive propensity-based weights that provide a consistency guarantee, relax overlap assumptions, and ideally reduce bias.

\end{itemize}

%% file: content/nontraditional.tex
\section{Beyond Traditional Data: Text, Networks, Images, and Treatment over Time}
\label{section:nontraditional}

As exciting as neural networks are for heterogeneous treatment effect estimation from quantitative data, a great promise of deep causal estimation is inference when treatments, confounders, and mediators are encoded in high-dimensional data (e.g., text, images, social networks, speech, and video) or are time-varying. This is a strong advantage of neural networks over other machine learning approaches, which do not generalize competitively to non-quantitative data.  In these scenarios, multi-task objectives and tailored architectures can be used to learn representations that are simultaneously rich, capture information about causal quantities, and disentangle their relationships. Moreover, the inherent flexibility of neural networks means that, in many cases, the TARNet-style models presented above can serve as the foundations to inference on text and graphs with some architectural modifications, additional losses, and new identification assumptions. 

This literature is rapidly evolving, so readers should treat this section of the primer as fundamentally prospective. To maintain accessibility, our primary goal here is to introduce readers to hypothetical scenarios where they might perform causal inference on text, network, or image data. Second, we selectively review contemporary, theoretically-motivated literature on deep causal estimation in these settings. The identification assumptions for different data types differ substantially, so we generally leave those to the interested reader. Finally, we briefly discuss approaches for dealing with time-varying confounding. We also take this section as an opportunity to introduce the Transformer or Graph Neural Network, an architecture now used in most contemporary deep learning models to learn from complex data (Box \ref{Box:gnn}).
 
\subsection{Causal Inference from Text}
\label{subsection:text}

In recent years, an interdisciplinary community across both social science and computer science has coalesced around causal inference from text (see \citet{keith-etal-2020-text} and \citet{feder2021causal} for exhaustive reviews). Broadly speaking, texts may capture information about any causal quantity (treatments, outcomes, confounders, mediators) we might be interested in. For example, in an exit-polling experiment, analysts might want to measure toxicity ($Y$) in text responses to political prompts. In an observational study of e-mail response times ($Y$), analysts might want to measure the effects of the tone of the email ($T$). In this scenario, the analyst might also want to control for confounders like subject matter ($X$). Each of these scenarios presents distinct identification challenges \citep{feder2021causal}. But in all cases, we can use low-dimensional representations of the high dimensional text to extract, quantify, and disentangle relationships between nuanced qualities like tone and subject matter. 

The ability of neural networks to automatically extract features makes them particularly suited for the last scenario when both treatment information and confounding covariates are encoded in text. In many cases, we may not have explicitly identified, quantified, or labeled all of the confounders in text (e.g., subject matter and tone of emails), but we would still like to control for them. \citet{pryzant2020causal},\citet{Veitch2019UsingTE}, and \citet{gui2022causal} address this problem by prepending Transformer-layers (Box \ref{Box:gnn}) for reading text to the beginning of TARNet or Dragonnet. \citet{Veitch2019UsingTE} demonstrate the viability of this approach on a Science of Science question testing the causal effect of equations on getting papers accepted to computer science conferences. \citet{pryzant2020causal,gui2022causal} explore the more complicated scenario not in which the treatment is explicitly known (e.g., equations in papers, gender of authors), but is instead externally perceived upon reading (e.g., politeness/rudeness of an email or toxicity of a social media post). In these models, an additional loss function is also added for learning text representations concurrently with the causal inference losses discussed above. 

 \begin{TCBBox}[label=Box:gnn]{Graph Neural Networks and Transformers}\
 \textbf{Graph neural networks} (GNNs) are the current state-of-the-art approach for creating representations for nodes in graphs. Compared to previous approaches that relied on ``shallow" embeddings based only on a node's local context (e.g., random walks to nearby nodes), GNNs are attractive because their node representations are aggregated from the structural position and covariates of all nodes $n$ degrees away from the target node, where $n$ is the number of graph neural network layers.

The most intuitive understanding of how graph neural networks work is as a message passing system \citep{gilmer2017neural}. We use one of the first GNN papers, the Graph Convolutional Network as an example \citep{kipf2016semi}. In this interpretation, each node has a message that it passes to it's neighbors through a graph convolution operation. In the first layer of a GNN this message would consist of the node's covariates/features. In consecutive layers of the network, these messages are actually representations of the node produced by the previous layer. During graph convolution, each node multiplies incoming messages by it's own set of weights and combines these weighted inputs using an aggregation function (e.g., summation). By the $n$-th GNN layer, these messages will contain structure and covariate information from all nodes $n$ degrees away. For interested readers, there is also a spectral interpretation of this process. Typically GNNs are trained to produce representations of graphs by predicting the probability that two nodes are linked in the network, and then used for something else. One variant of the GNN uses an ``attention" mechanism to vary the extent that nodes value messages from different neighbors (the graph attention network or GAT) \citep{velivckovic2017graph}. 

As of 2023, \textbf{Transformers} are the hegemonic architecture used in natural language processing. After their introduction in 2017, these models improved performance on many high-profile NLP tasks across the board. Several enterprise-scale transformers have been featured in the media for their impressive performance in text generation and question answering (e.g. OpenAI's GPT-3 and ChatGPT, Google's Bard). Smaller models in broad use are based on the BERT architecture \citep{devlin2018bert}.

The connection between GNNs, and specifically GATs, is the focus on attention mechanisms. From this perspective, words in sentences are akin to nodes in networks, with their relative positions to each other being analogous to their structural positions in the graph. Transformers improved on previous sequential approaches to text analysis (i.e. recurrent neural networks) by having each word (or representation of a word) receive messages from not just adjacent words, but all words heterogeneously. Attention mechanisms throughout the architecture allow each layer of a transformer to attend to words or aggregated representation mechanisms heterogeneously. Architectures such as BERT or GPT stack transformer layers to create models with hundreds of millions of parameters. These models are expensive to train, both computationally and with respect to data, so they are often pretrained on large datasets and then ``fine-tuned" (lightly re-trained) with smaller datasets for specific tasks or to align with certain goals.  

\end{TCBBox}

\subsection{Causal Inference from Networks}
\label{subsection:networks}

A smaller literature has leveraged relational data for causal inference in two distinct scenarios. In the first traditional selection on observable settings, we wish to control for information about unobserved confounding inferable from homophilous ties. For example, age or gender might be unmeasured in our data, but we might expect people to develop friendship ties with those of the same gender identity or age cohort.

This scenario suggests estimation strategies similar to those when confounders are encoded in text. Much like Transformer layers can be prepended to TARNet-style estimators to learn from text, graph neural networks (an analog of the Transformer) can be preprended to learn from graphs. \citet{guo2019counterfactual} provides a first pass at this problem by adding  GNN layers to CFRNet \cite{Shalit2017}. \citet{Veitch2019UsingET} instead adapt Dragonnet in a semi-parametric framework to allow for consistent estimates of the treatment and outcome, assuming the network representation encodes significant information about confounders.

The second, more challenging scenario is estimating the causal effect of social influence on outcomes from observational data. For example, \citet{Cristali2022UsingEF} introduce the problem of measuring the effects of vaccination ($T$) on peer vaccination choice ($Y$). This is a hard problem because a) SUTVA is a fundamental assumption of all causal inference frameworks and b) it is hard to disentangle whether changes in the outcome result from the treatment via peer effects (e.g, person A pressuring person B to vaccinate), or from homophily (e.g., person A and person B having similar political leanings). In other words, contagion and homophily are generically confounded \citep{shalizithomas2011}. \citet{mcfowland2021} are the first to tackle this problem by making strong parametric assumptions about the generation of network ties and the outcome model. \citet{Cristali2022UsingEF} instead propose an approach using neural network-learned representations of the graph.     

\subsection{Causal Inference from Images}
\label{subsection:images}

While ideas from causal inference have been leveraged extensively to improve image classification, to our knowledge there are no papers that explore causal inference where treatments, confounders, mediators, or predictors are encoded in images.\footnote{\citet{jesson2021quantifying} introduce a simulation where the MNIST digit dataset serves as covariates $X$ as toy example of high-dimensional confounding, but not a possible application.} That being said, some scenarios proposed for causal text analysis should apply here as well. For example, consider the conjoint experiment by \citet{todorov2005inferences} where both the treatment (e.g. incumbency of a politician) and potential latent confounders (e.g., party, age, gender, race) are encoded in an image. In this setting, a TARNet-like model adapted to learn and condition on image representations could improve treatment effect estimation by controlling for confounders such as the politician's age. Causal inference on images is an area ripe for exploration, and we hope to see more work here in the future.

\subsection{Causal Inference from Time-varying Data}
\label{subsection:timevarying}

One natural extension of deep causal estimation is to scenarios where treatments are administered over time and confounding may be time-varying. While “g-methods” developed by Robins et al. for estimating effects with time-varying treatments and confounding have existed for decades, the statistical assumptions encoded in these models are quite strong \citep{robins1994correcting,robins2000marginal,robins2009longitudinal}. Due to their reliance on generalized linear models to define the “structural” component, they assume that the outcome is a linear function of all covariates and treatment. Second, for identification, they make strong assumptions about which previous timesteps confound the current one. Third, they require different coefficients to be estimated at each time steps. Transformers (Box \ref{Box:gnn}) and recurrent neural networks, a simpler model for sequential data (Appendix A.\ref{appendix:rnn}), should be able to capture long-term dependencies and non-linearities in ways that marginal structural models and g-computation cannot.

Several papers have begun to explore these possibilities in the context of personalized medicine. \citet{lim2018forecasting} build a marginal structural model using a recurrent neural network, and \citet{bica2020estimating} extend this framework with an additional loss to more explicitly deal with time varying confounding by forcing the model to “unlearn” information about the previous time steps. \citet{melnychuk2022} go one step further by adapting \citet{bica2020estimating}’s approach with a transformer. Inspired by longitudinal targeted maximum likelihood, \citet{frauen2022estimating} add a semi-parametric targeting layer to their RNN to create a g-computation algorithm that is doubly robust and asymptotically efficient. \citet{lietal2021} instead propose an RNN framework for g-computation that allows for dynamic treatment regimes. All of these papers use simulations of tumor growth dynamics, naturalistic simulations based on vital signs from intensive care unit visits, or factual datasets exploring treatment response to physical therapy for back pain.

%% file: content/conclusion.tex
\section{Conclusion: Deep Causal Estimation in Context}
\label{section:conclusions}

In this primer we introduce social scientists to the emerging machine learning literature on deep learning for causal inference. To set the stage, we first provide both an intuitive introduction to fundamental deep learning concepts like representation and multi-task learning, as well as practical guidelines for training neural networks. In the main body of the article, we show how ML researchers have adapted core treatment and outcome modeling strategies to leverage the particular strengths of neural networks for heterogeneous treatment effect estimation. We follow with a discussion on inference (e.g., model selection, confidence intervals, interpretation), and closed with a prospective look at algorithms for inference from text, social networks, images, and time varying data.

Deep learning is not the only potential tool for heterogeneous treatment effect inference, and there are robust literatures exploring the usage of other methods in both the econometrics and biostatistics communities \citep{van2011targeted,chernozhukov2016double,wager2018}. While these literatures are certainly more mature, below we discuss reasons why we think the use gap between neural networks and other machine learning methods will continue to narrow, a change that we must prepare for.
 
First, neural networks are better at modeling non-linear heterogeneity (e.g., in treatment responses) than other machine learning methods. In extensive simulations, \citet{curth2021} found that when the data-generating process for treatment heterogeneity includes exponential relationships, neural networks outperformed random forests, but tree-based methods are robust when the data-generating process is built on linear functions. Neural networks were also consistently better at predicting outlier treatment effects than forests. These differences result from how the two methods model functions. While neural networks can approximate any continuous function with enough neurons, random forests must build non-linear or non-orthogonal decision boundaries using piecewise functions and average predictions. Consistent with these differences, \citet{curth2021} also find that neural networks do better when variables are constructed as continuous covariates, and vice versa when they are dichotomized.  

From a statistical perspective, the rise of semi-parametric and double machine learning frameworks has also narrowed the gap between neural networks and other types of machine learning in terms of theoretical guarantees. For example, the TMLE-inspired Dragonnet algorithm featured here is unbiased, plausibly consistent, and converges to the target estimand at a fast rate of $\frac{1}{\sqrt{n}}$. The closely-related Riezsnet double machine learning model (not featured) boasts similar guarantees \citep{chernozhukov2022riesznet}. Beyond these algorithms, there is a growing adjacent literature of model-agnostic plug-in learners (e.g., X-learner, R-learner) that can leverage the strengths of neural networks \citep{nie2021quasi,kunzel2019metalearners}.

Third, folk beliefs about the data-hungriness and uninterpretability of neural networks are overstated. Neural networks are data-hungry when over-parameterized or learning from high-dimensional data like images, but we show in the tutorials that modest-sized, well-regularized neural networks can successfully infer heterogeneous treatment effects in a naturalistic simulation of quantitative data with less than 800 units. In Section \ref{section:interpretation}, we also highlight the considerable progress in machine learning interpretability over the past five years, much of which has been on model-agnostic approaches that benefit all black-box algorithms equally.\footnote{Critics often point to out-of-bag feature importances as a particular strength of random forests, but this approach has been shown to be less accurate than model-agnostic permutation importances anyways \citep{altmann2010}.}

In our opinion, the most pressing limitation of current deep learning approaches is the difficulty of optimizing neural networks. Theoretically, this stems from a) the complexity of the loss functions which are often non-convex, and b) the ease of over-parameterizing these models to fit these functions. If neural networks are to be used as statistical estimators, statistical guarantees must be backed by optimization guarantees and/or more rigorous methods for model selection. Outside of statistical estimation, this limitation has largely been addressed through empirical testing on test data and strategic model selection. Within the statistical estimation context, this gap will likely need to be addressed by simulation-based sensitivity analyses and, in the short term, comparisons to other model families.

Moreover, there has been a lack of mature tools and empirical applications of these models. A major goal of this primer, and the tutorials in particular, is to synthesize the theoretical literature, practical training and interpretation guidelines,  and annotated code so that social scientists in one place can start using these models. Deep learning frameworks like Tensorflow and Pytorch are becoming more accessible every year, but we note that canned Python packages like Uber's causalML exist for interested readers who just want to experiment with a few of these models \citep{chen2020causalml}.
 
Despite current limitations, we believe the future of causal estimation runs through deep learning. As causal inference ventures into new settings, the flexibility of neural networks will become essential for learning from text, graph, image, video, and speech data. For time-varying settings, we believe the ability of neural networks to model non-linearities and long-range temporal dependencies will ultimately lead to solutions with net weaker assumptions than current approaches. Overall, we are optimistic and excited to see where deep causal estimation heads over the next few years.

\section{Author's Note}
The accompanying tutorials are available at \href{https://github.com/kochbj/Deep-Learning-for-Causal-Inference}{https://github.com/kochbj/Deep-Learning-for-Causal-Inference}. The tutorials use the IHDP naturalistic simulation introduced in \citet{Hill2011} as an example. The  25 covariates/features for the 747 units (139 treated) in the dataset were taken from an experiment, but Hill simulated the outcomes to create known counterfactuals. The data are available from Fredrik Johansson's website \href{https://www.fredjo.com}{https://www.fredjo.com}.

%% file: content/appendix.tex
\section{Balancing Using Integral Probability Metrics}
\label{appendix:cfrnet}
\subsection{Wasserstein Distance}
Following \citep{stockblog, dazablog}, suppose we have two discrete distributions (treated and control) with marginal densities $p(x)$ and $q(x)$ captured as vectors $t$ and $c$, with dimensions $n$ and $m$ respectively. To compute the Wasserstein distance, we must define a "mapping matrix" $P$ that defines the mapping of ``earth" in $p(x)$ to corresponding piles in $q(x)$. Let $\textbf{U}(t,c)$ be the set of positive, $n \times m$   mapping matrices where the sum of the rows is $t$ and the sum of the columns is $c$.  
\begin{equation}
\textbf{U}(t,c)= P\subset \mathbb{R}^{nxm}_{>0}| P \cdot 1_m=\textbf{t} , P^T \cdot 1_n=\textbf{c}
\end{equation} 
In words, this matrix maps the probability mass from points in the support of $p(x)$ (i.e, the elements of $t$) to points in the support of $q(x)$ (the elements of $c$) (note that the mapping need not be one-to-one). We also have a ``cost" matrix $C$ that describes the cost of applying $P$ (i.e. the cost of shoveling dirt according to the map described in $P$). The cost matrix can be computed using a norm $\ell$ (most commonly $\ell^2$) between the points in $t$ being mapped to $c$ in the mapping matrix $P$. Finally, the $\ell$-norm Wasserstein distance $dW_\ell$ can be defined as 
\begin{equation}
dW_\ell = min_{P\subset(t,c)}\sum_{i,j}P_{ij}C_{ij}
\end{equation}
In other words, the Wasserstein distance is the smallest Frobenius inner product of a mapping matrix $P$ that fits the above constraints, and its associated cost matrix $C$. Although this problem can be solved via linear programming, the Wasserstein distance is often implemented in a different form that works with continuous distributions and can be optimized by gradient descent \citep{arjovsky2017wasserstein,gulrajani2017improved}. There is also a variant of the Wasserstein distance that imposes an entropy-based regularization on the coupling matrix to make it smoother or sparser called the Sinkhorn distance \citep{cuturi2013sinkhorn}.

\subsection{Extending Representation Balancing with IPMs}
\textbf{Deep Dive: CFRNet (\citet{Shalit2017,Johansson2018,johannson2020})}

Beyond receiving outcome modeling gradients for both potential outcomes, the authors have subsequently extended TARNet with additional losses that  explicitly encourage balancing by minimizing a statistical distance between the two covariate distributions in representation space. These distances are called integral probability metrics \citep{muller1997integral}.\footnote{\citet{zhang2020learning} criticize the usage of IPMs because they make no restrictions on the moments of the transformed distributions. Thus while the covariate distributions may have a high percentage of overlap in representation space, this overlap may be substantially biased in unknown ways.} \citet{Johansson2016,Shalit2017,Johansson2018} propose two possible IPMs, the Wasserstein distance and the maximum mean discrepancy distance (MMD) for use in these architectures. 

The Wasserstein or ``Earth Mover's" distance fits an interpretable ``map" (i.e. a matrix) showing how to efficiently convert from one probability mass distribution to another.  The Wasserstein distance is most easily understood as an optimal transport problem (i.e., a scenario where we want to transport one distribution to another at minimum cost). The nickname ``Earth mover's distance" comes from the metaphor of shoveling dirt to terraform one landscape into another. In the idealized case in which one distribution can be perfectly transformed into another, the Wasserstein map corresponds exactly to a perfect one-to-one matching on covariates strategy \citep{kallus2016generalized}.

The MMD is the normed distance between the means of two distributions, after a kernel function $\phi$ has transformed them into a high-dimensional space called a reproducing kernel Hibbert Space (RKHS) \citep{gretton2012kernel}. The MMD with an $L^2$ norm in RKHS $\mathcal{H}$ can be specified as:

\begin{equation}
MMD(P,Q) = ||\Ex_{X \sim P}\phi(X) - \Ex_{X \sim Q}\phi(X)||^2_{\mathcal{H}}
\end{equation}

The metric is built on the idea that there is no function that would have differing Expected Values for $P$ and $Q$ in this high-dimensional space if $P$ and $Q$ are the same distribution \citep{huszarblog}. The MMD is inexpensive to calculate using the ``kernel trick" where the inner product between two points can be calculated in the RKHS without first transforming each point into the RKHS.\footnote{This kernel trick is  also what makes support vector machines computationally tractable.} 

\begin{figure}
    \centering
    \includegraphics[width=\linewidth]{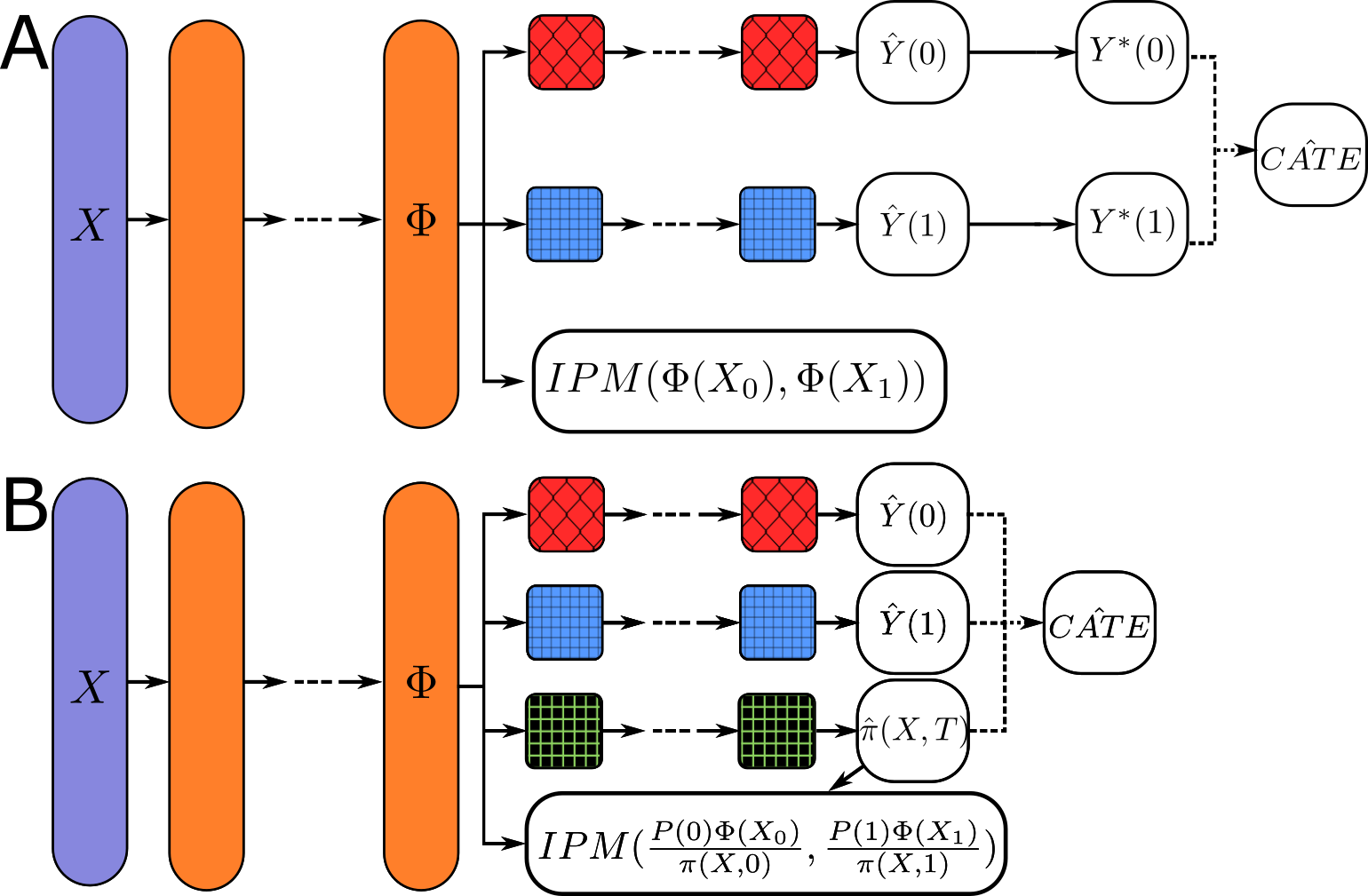}
    \caption{\textbf{A. CFRNet} architecture originally introduced in \citet{Shalit2017}. CFRNet adds an additional integral probability metric (IPM) loss to  TARNet to explicitly force representations of the treated and control covariates closer in representation space.
    \\\textbf{B. Weighted CFRNet} adds a propensity score head to CFRNet to predict IPW-weighted outcomes. During training, the propensity score is used to reweight both the predicted outcomes $\hat{Y}(0)$ and $\hat{Y}(1)$, as well as the represented covariate distributions in calculation of the IPM loss. This allows the authors to provide consistency guarantees and relax the overlap assumption. Figures adapted from \citet{johannson2020}.}
    \label{fig:cfrnet}
\end{figure}

When an IPM loss is applied to the representation layers in TARNet, the authors call the resulting network ``CounterFactual Regression Network" (CFRNet) (Fig. \ref{fig:cfrnet}A) \citep{Shalit2017}. The loss function for this network is

\begin{equation}
\min_{h,\Phi,IPM}\frac{1}{N}\sum_{i=1}^N \underbrace{MSE(Y_i,h(\Phi(X_i),T_i))}_{\text{Outcome Loss}} +\lambda\underbrace{IPM(\Phi(X|T=1),\Phi(X|T=0))}_{\text{Dist. b/w T \& C covar. distributions}} +  \alpha\underbrace{\mathcal{R}(h)}_{L_2} 
\end{equation}
where $\mathcal{R}(h)$ is a model complexity term and $\lambda$ and $\alpha$ are hyperparameters.

These two papers also make important theoretical contributions by providing bounds on the generalization error for the PEHE \citep{Hill2011}. In \citet{Shalit2017}, they show that the PEHE is bounded by the sum of the factual loss, counterfactual loss, and the variance of the conditional outcome.

In \citet{johannson2020}, the authors introduce estimated IPW weights $\pi(\Phi(X),T)$ to CFRNet that are used within the IPM calculation to provide consistency guarantees (Fig. \ref{fig:cfrnet}B). Theoretically, they also use these weights to relax the overlap assumption as long as the weights themselves obey the positivity assumption. From a practical standpoint, adding weights that are optimized smoothly across the whole dataset each epoch reduces noise created by calculating the IPM score in small batches. Weighted CFRNet minimizes the following loss function:

\begin{equation}
\begin{split}
\argmin_{h,\Phi,IPM,\pi,\lambda_h,\lambda_w}\frac{1}{N}\sum_{i=1}^N \underbrace{\frac{\hat{P}(T_i)}{\pi(\Phi(X_i),T_i)}}_{IPW}\cdot \underbrace{MSE(Y_i,h(\Phi(X_i),T_i))}_{\text{Outcome Loss}} + \lambda_h \underbrace{\mathcal{R}(h)}_{\text{$L_2$ Outcome}}+\\
\alpha\cdot \underbrace{IPM(\underbrace{\frac{\hat{P}(1)}{\pi(\Phi(X,1))}}_{IPW}\cdot\Phi(X|T=1),\underbrace{\frac{\hat{P}(0)}{\pi(\Phi(X,0))}}_{IPW}\cdot\Phi(X|T=0))}_{\text{Distance between $IPW$ weighted T \& C covar. distributions}}+\lambda_\pi\underbrace{\frac{||\pi||_2}{N}}_{L_2   Var(\pi)}
\end{split}
\end{equation}
where $\mathcal{R}(h)$ is a model complexity term and $\lambda_h$, $\lambda_\pi$ and $\alpha$ are hyperparameters. The final term is a regularization term on the variance of the weight parameters.

\subsubsection{Extending Representation Balancing with Matching}
Beyond IPMs, other approaches have directly embraced matching as a balancing strategy. \citet{Yao2018} train their TARNet on six point mini-batches of propensity score-matched units with additional reconstruction losses designed to preserve the relative distances between these points when projecting them into representation space. \citet{Schwab2018} takes an even simpler approach by feeding random batches of propensity-matched units to the TarNet outcome structure.

\section{Model Selection Using the PEHE}
\label{appendix:peheselection}
In order to select hyperparameters in real data, \citet{johannson2020} propose to use a matching variant of $PEHE$ with the nearest Euclidean neighbor of each unit $i$ from the other treatment assignment group $y_i^{nn}$ as a counterfactual. If we identify the nearest neighbor $j$ of each unit $i$ in representation space such that $t_j\neq t_i$ as

  $$y_i^{nn}(1-t_i) = \min_{j\in (1-T)}||\Phi(x_i|t_i)-\Phi(x_j|1-t_i)||_2$$
 then,
$$PEHE_{nn}=\frac{1}{N}\sum_{i=1}^N{(\underbrace{(1-2t_i)(y_i(t_i)-y_i^{nn}(1-t_i)}_{CATE_{nn}}-\underbrace{(h(\Phi(x),1)-h(\Phi(x),0)))}_{\hat{CATE}}}^2$$
If we take the square root of the $PEHE_{nn}$ then we get an approximation of the unit-level error.

The intuition behind $\sqrt{PEHE_{nn}}$ is solid. If our representation function $\Phi$ is truly learning to balance the treated and control distributions, $CATE_{nn}$ should coarsely measure it.

\section{Recurrent Neural Networks (RNN)}
\label{appendix:rnn}

\begin{figure}
\centering
\includegraphics[width=.5\textwidth]{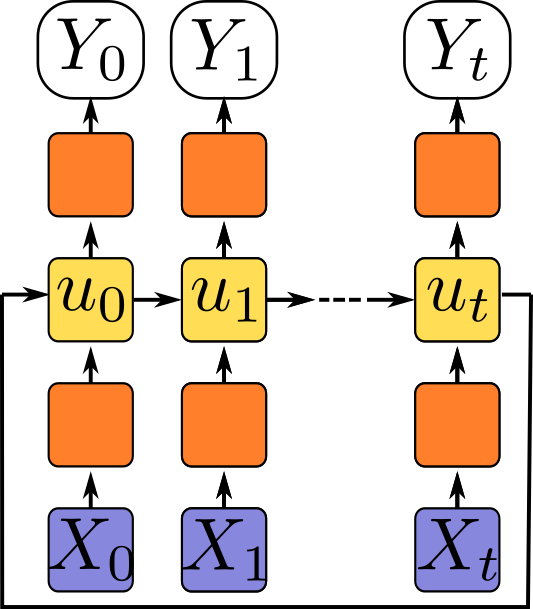}
\caption{Recurrent neural network.}
\label{fig:rnn}
\end{figure}

Recurrent neural networks are a specialized architecture created for learning outcomes from sequential data (e.g. time series, biological sequences, text) (Fig. \ref{fig:rnn}). In a classic RNN, each ``unit" $u$ in the network takes as input its own covariates $X$ (or possibly a representation) and a representation produced by the previous unit, encoding cumulative information about earlier states in the sequence. These units are not just simple hidden layers: there is a set of weights within each unit for its raw inputs, the representation from the previous time step, and its outputs. Different RNN variants have different operations for integrating past representations with present inputs. Recurrent neural networks may be directed acyclic graphs or feedback on themselves. Commonly used variants include Gated Recurrent Unit networks (GRU) and Long-term Short-term memory networks (LSTM) \citep{cho2014gru,lstm1997}.

\section{Generative Modeling through Adversarial Training}
\label{appendix:generative}
Adversarial training approaches include a wide variety of architectures where two networks or loss functions compete against each other. Adversarial approaches are inspired by Generative Adversarial Networks (GANs) (Box \ref{box:gan}) \citep{goodfellow2014generative}. In the machine learning literature on causal inference, adversarial training has been applied both to trade off outcome modeling and treatment modeling tasks during representation learning, as well as to trade off estimation and regularization of IPW weights. GANs have also been used directly as generative models for counterfactual and treatment effect distributions. 

\begin{TCBBox}[label=box:gan]{Generative Adversarial Networks (GAN)}In GANs, two networks, a discriminator network $D$ and a generator network $G$, play a zero-sum game like cops and robbers. The generator network's job is to learn a distribution from which the training data $X$ could have credibly been generated. In each training batch, the generator produces a new outcome (originally images, but could be IPW weights, counterfactuals or treatment effects) by drawing a random noise sample from a known distribution $Z$ (e.g. Gaussian) and transforming it into outcomes with the function $G(Z)=\hat{X}$. The discriminator's job is to learn a function $D(X)=P(X\ is\ real)$ that can distinguish whether the outcome is from the training data $X$, or whether it is a ``fake" $\hat{X}$ created by the generator. The generator then receives a negative version of the discriminator's loss, a penalty that is proportional to how well it was able to ``deceive" the discriminator.  The discriminator's loss can be the log loss,  Jensen-Shannon divergence \citep{goodfellow2014generative}, the Wasserstein distance \citep{arjovsky2017wasserstein, gulrajani2017improved}, or any number of divergences and IPMs. Formally, the generator attempts to minimize the following loss function,
\begin{equation*}
\argmin_{G}=\underbrace{\Ex_{X}}_{\text{ real dist.}}[\underbrace{\mathcal{L}(D(X)}_{P(X\ is\ real)}] + \underbrace{\Ex_{Z}}_{\text{fake dist.}}[1-\underbrace{\mathcal{L}(D(G(Z))}_{P(\hat{X}\text{is real})}]
\end{equation*}
where the first quantity is the discriminator's estimated probability data from $X$ is indeed real, and the second quantity is the discriminator's estimate that a generated quantity from the distribution $Z$ is real. 

Because the discriminator is trying to catch the generator, its objective is to maximize the same function,
\begin{equation*}
\argmax_{D}=\underbrace{\Ex_{X}}_{\text{ real dist.}}[\underbrace{\mathcal{L}(D(X)}_{P(X \text{is real})}] + \underbrace{\Ex_{Z}}_{\text{fake dist.}}[1-\underbrace{\mathcal{L}(D(G(Z))}_{P(\hat{X}\text{is real})}]
\end{equation*}
In practice, the discriminator and the generator are trained either alternatingly or simultaneously, with the discriminator increasing its ability to discern between real and fake outcomes over time, and the generator increasing its ability to deceive the discriminator. The idea is that the adaptive loss function created by the discriminator can coax the generator out of local minima to generate superior outcomes. Results by these models have been impressive, and many of the fake portraits and ``deepfake" videos circulating online in recent years are generated by this architectures. The advantage of GANs is that they can impressively learn very complex generative distributions with limited modeling assumptions. The disadvantage of GANs is that they are difficult and unreliable to train, often plateauing in local optima.
\end{TCBBox}
\label{Box:GANs}

\subsection{GANs as Generative Models of Treatment Effect Distributions (GANITE)}
\begin{figure}
    \centering
    \includegraphics[width=\linewidth]{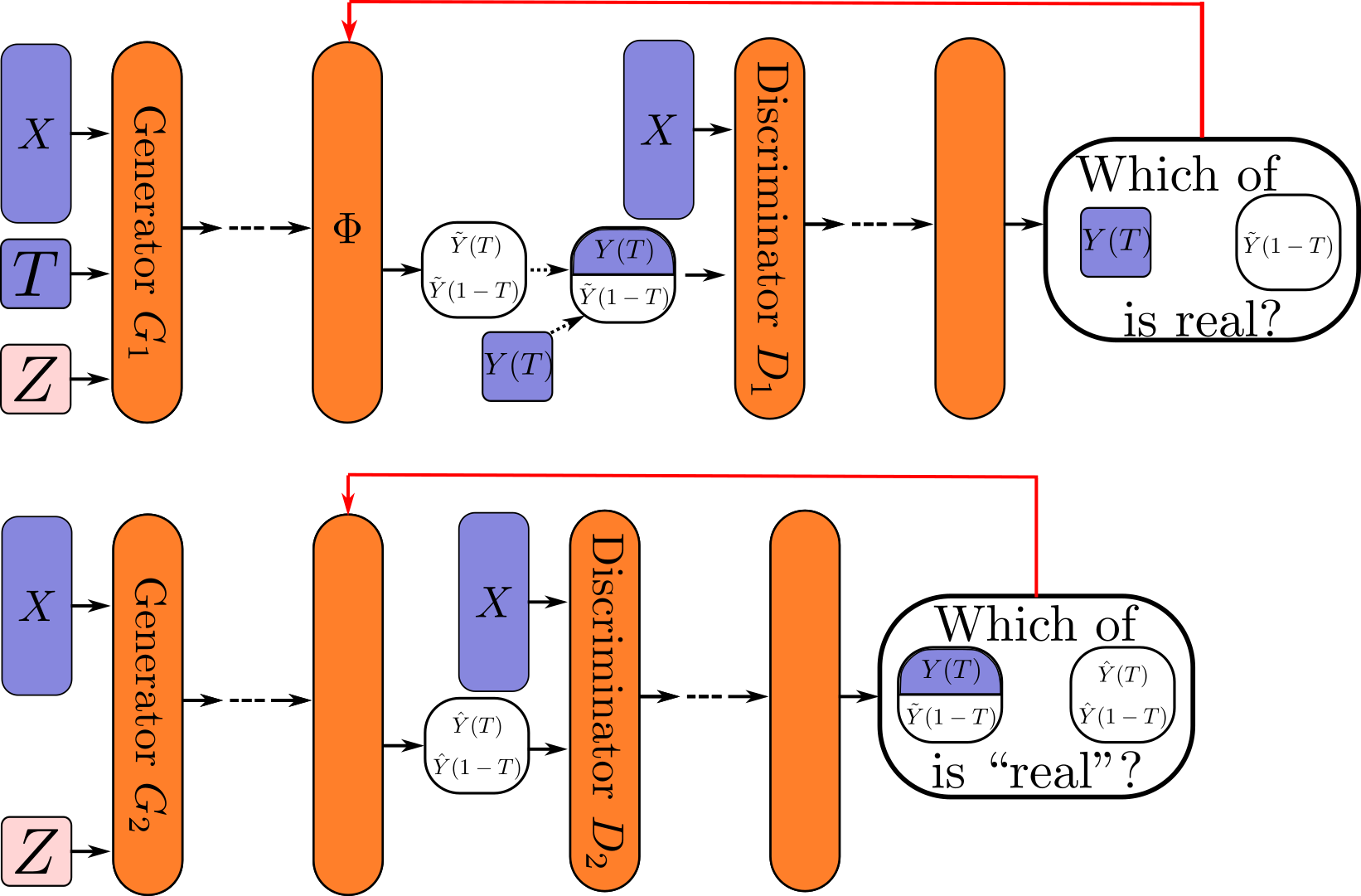}
    \caption{\textbf{GANITE} has two GANs. The first generator $G_1$ generates counterfactuals $\tilde{Y}(T)$. The discriminator $D_1$ attempts to discriminate between these predictions and real data ($Y(T)$). The second generator proposes pairs of potential outcomes $\hat{Y}(0)$ and $\hat{Y}(1)$ (i.e., treatment effects), a vector we call $R$. Discriminator $D_2$ attempts to discern between $R$ and a ``complete dataset" $C$ created by pairing each observed/factual outcome $Y(T)$ with a synthetic outcome $\tilde{Y}(1-T)$ proposed by $G_1$. Although we do not show gradients in other figures, we make an exception for GANs (solid red line).}
    \label{fig:ganite}
\end{figure}

\textbf{Deep Dive: GANITE (\citet{Yoon2018})}
Although a generative model of the treatment effect distribution is generally unknown, a natural application of GANs is to try to machine learn this distribution from data. GANITE uses two GANs: $GAN_1$, consisting of generator $G_1$ and discriminator $D_1$, to model the counterfactual distribution and $GAN_2$, consisting of generator $G_2$ and discriminator $D_2$, to model the $CATE$ distribution \citep{Yoon2018} (Fig. \ref{fig:ganite}).
The training procedure for $GAN_1$ is as follows:
\begin{enumerate}
\singlespacing
    \item Taking $X$,$T$, and generative noise $Z$ as input, generator $G_1$ generates both potential outcomes $\{\tilde{Y}(T),\tilde{Y}(1-T)\}$. A factual loss $MSE(Y(T),\tilde{Y}(T))$ is applied.
    \item Create a new vector $\textbf{C}=\{Y(T),\tilde{Y}(1-T)\}$ by combining the observed potential outcome and the counterfactual predicted by $G_1$.
    \item Taking $X$ and $\textbf{C}$ as inputs, the discriminator $D_1$ rates each value in $\textbf{C}$ for the probability that it is the observed outcome using the categorical cross entropy loss:
    \begin{equation}
    \mathcal{L}(D_1)=CCE(\{\underbrace{P(\textbf{C}_0=Y(T))}_{\text{Prob first idx is real}},\underbrace{P(\textbf{C}_1=Y(T))}_{\text{Prob sec idx is real}}\},\{\underbrace{\textbf{C}_0==Y(T)}_{\text{1 if idx 0 is real}},\underbrace{\textbf{C}_1==Y(T)}_{\text{1 if idx 1 is real}}\})
\end{equation}
    \item This loss is then fed back to $G_1$ such that the total loss for the generator is now  
    \begin{equation}
        \argmin_{G_1}=MSE(Y(T),\tilde{Y}(T)) - \lambda \mathcal{L}(D_1)
    \end{equation}
\end{enumerate}

After generator $G_1$ is trained to completion, the authors use $\textbf{C}$ as a ``complete dataset" containing both a factual outcome and a counterfactual outcome to train $GAN_2$, which generates treatment effects: 
\begin{enumerate}
\singlespacing
    \item Taking only $X$ and generative noise $Z$ as input, $G_2$ generates a new potential outcome vector $\textbf{R}=\{\hat{Y}(T),\hat{Y}(1-T)\}$. $G_2$ receives an MSE loss to minimize the difference between its predictions and the ``complete dataset" $\textbf{C}$: $MSE(\textbf{C},\textbf{R})$.
    \item Discriminator $D_2$ takes $X$, $\textbf{C}$, and $\textbf{R}$ as inputs and estimates a probability that $\textbf{C}$ is the ``complete" dataset, and that $\textbf{R}$ is the ``complete dataset":
    \begin{equation}
    \mathcal{L}(D_2)=CCE(\{\underbrace{P(\textbf{C}=\textbf{C})}_{\text{ Prob $\textbf{C}$ is ``CD" }},\underbrace{P(\textbf{R}=\textbf{C})}_{\text{Prob $\textbf{R}$ is ``CD"}}\},\{\underbrace{\textbf{C}==\textbf{C}}_{\text{1 if idx 0 is $\textbf{C}$}},\underbrace{\textbf{C}_1==Y(T)}_{\text{1 if idx 1 is $\textbf{C}$}}\})
\end{equation}
    \item This loss is then fed back to the generator $G_2$ such that the total loss for the generator is now  
    \begin{equation}
        \argmin_{G_2}=MSE(\textbf{C},\textbf{R}) - \lambda \mathcal{L}(D_2)
    \end{equation}
\end{enumerate}
At the end of training, $G_2$ should be able to predict treatment effects with only covariates $X$ and noise $Z$ as inputs. An evolution of GANITE, SCIGAN, extends this framework to settings with more than one treatment and continuous dosages \citep{bica_sicgan}. 

\newpage
\subsection{Adversarial Representation Balancing}

The use of the IPM loss in CFRNet \citep{Shalit2017} may also be viewed as an adversarial approach in that the representation layers are forced to maximize performance on two competing tasks: predicting outcomes and minimizing an IPM. Rather than using an IPM loss, other authors have trained propensity score estimators that send positive (rather than negative) gradients back to the representation layers \citep{Atan2018,Du2019}.

\citet{bica2020estimating} extend this approach to settings with treatment over time using a recurrent neural network. In their medical setting, decorrelating treatment from patient covariates and history allows them to estimate treatment effects at each individual snapshot.